\documentclass[twocolumn]{article}          
\usepackage{hyperref}
\usepackage[dvips]{graphicx}
\usepackage[latin1]{inputenc}
\usepackage{amssymb,amsmath,nccmath}
\usepackage{subfig}

\usepackage{graphics}
\usepackage{epsfig}
\usepackage{color}
\usepackage{multirow} 
\usepackage{algorithm}
\usepackage{algpseudocode}
 
\newcommand{\RR}[1]{\mathbb{R}^{#1}} 
\newcommand{\bd}[1]{\mbox{\boldmath $#1$}}

\newcommand{\mbs}[1]{\boldsymbol{#1}}

\newcommand{\mc}[1]{\mathcal{#1}}

\begin{document}

\date{\color{red}{Received: 29 April 2014 / Accepted: 7 June 2015} \\
published in PAAA}

\title{Approximate Variational Inference Based on a Finite Sample of Gaussian Latent Variables}

\author{Nikolaos Gianniotis \and Christoph Schn\"orr \and Christian Molkenthin \and Sanjay Singh Bora}


\maketitle

\begin{abstract}

Variational methods are employed in situations where exact Bayesian inference
becomes intractable due to the difficulty in performing certain integrals.
Typically, variational methods postulate a tractable posterior and formulate a lower bound 
on the desired integral to be approximated, e.g. marginal likelihood.
The lower bound is then optimised with respect to its free parameters, the so called variational parameters.
However, this is not always possible as for certain integrals it is very challenging (or tedious)
to come up with a suitable lower bound.
Here we propose a simple scheme that overcomes some of the awkward cases where the usual variational
treatment becomes difficult. The scheme relies on a rewriting of the lower bound on the model log-likelihood.
We demonstrate the proposed scheme on a number of synthetic and real examples, 
as well as on a real geophysical model for which the standard variational approaches are inapplicable.
\end{abstract}

\section{Introduction}
\label{sec:introduction}

Bayesian inference is becoming the standard mode of inference as
computational resources increase, algorithms advance and scientists
across fields become aware of the importance of uncertainty.
However, exact Bayesian inference is hardly ever possible whenever
the model likelihood function deviates from mathematically convenient
forms (i.e. conjugacy). Deterministic approximations are constantly
gaining ground on the ubiquitous and computationally intensive Monte
Carlo sampling methods that are capable of producing high quality 
approximations to otherwise intractable quantities such as 
posterior densities or marginal likelihoods.

Often, however, deterministic schemes are tailored to
a particular model, or family of models, and hence previously
derived  methods might not be transferable to
a new setting (e.g. in \cite{Barber1998} a specialised algorithm for Bayesian inference in neural networks is considered).
 This introduces practical difficulties, mostly in fields
beyond machine learning, whenever an implementation of Bayesian
inference is required particularly in early explorative stages
where the model formulation is likely to keep changing.
In this work we introduce a scheme for approximate
Bayesian inference that aims to be general enough so that
it can accommodate a variety of models. 
The proposed scheme is conceptually simple and requires only
that the gradient of the model log-likelihood with respect to its parameters
be available.

Specifically, we consider the task of computing a global Gaussian approximation
$q(\bd{w}) = \mc{N}(\bd{w}|\mbs\mu,\mbs\Sigma)$ to a given intractable posterior
distribution representing a probabilistic model by maximizing a standard variational
lower bound \cite{VariationalGaussianApprox-09}. This lower bound involves the 
expectation of the log-likelihood of the model distribution with respect to the 
approximating distribution $q(\bd{w})$. To enable the computation of these expectations
either in closed form or through computationally tractable numerical approximations, the
likelihoods are typically restricted to conditionally factorized forms. This step of the
approach specifically depends on the model at hand.

By contrast, our conceptually simple method presented below is more generally applicable
to various models in the same way. We demonstrate this by working out examples for a variety of
models. In particular, we demonstrate empirically that our approach results in
Gaussian approximations that are superior to the basic Laplace approximation \cite{Bishop}, which is the
typical objective of variational approximation schemes. A formal comparison to related 
state-of-the-art methods for computing improved Gaussian approximations, e.g.~through 
the nested Laplace approximation \cite{FastGMRFLatent-08} or by expectation propagation
\cite{EP-MarginalsGaussian-11}, is beyond the scope of this paper, however.

Our paper is organized as follows: in Section \ref{sec:proposed_scheme} we introduce
in general terms the proposed scheme. Starting from a standard formulation,
that typically variational methods use, we postulate a Gaussian posterior $q(\bd{w})$
and show how to form an approximation
to the lower bound of the marginal log-likelihood. The obtained approximation
allows us to optimise the variational parameters $\mbs\mu$ and $\mbs\Sigma$ of $q(\bd{w})$ by
gradient optimisation. For the reader that wishes to refresh her/his memory or obtain
a more detailed explanation of the equations presented in Section \ref{sec:proposed_scheme}, we refer to \cite{Beal2003,Tzikas2008}.

In Section \ref{sec:applications} we demonstrate
the proposed scheme on a number of applications and 
compare it against exact inference, Laplace and variational approximations.
Specifically, in Subsection \ref{sec:fitting_bivariate} we show a visual
comparison of approximating flexible bivariate densities using the Laplace approximation and the proposed scheme.
In Subsection \ref{sec:blr} we apply our approach on the problem of Bayesian 
linear regression which actually does admit an analytical and exact solution. This
is useful as it allows us to empirically verify the correctness of our scheme
against the posterior obtained by exact inference.
Subsequently, in Subsections \ref{sec:logistic_regression}
and \ref{sec:multiclass} we compare  the proposed scheme with approaches
that take into account the functional forms of classification problems.
We show that despite its general formulation, the proposed scheme performs up to par in this setting
without exploiting any such problem specific knowledge. In Subsection \ref{sec:denoising} we show
how a change in the model likelihood of probabilistic principal component analysis \cite{Tipping1999},
that renders inference problematic, can easily be accommodated by the proposed scheme. This demonstrates
the versatility of our approach in handling such cases in a direct and simple manner. Finally, 
in Subsection \ref{sec:psha} we show how the proposed scheme can be applied beyond the usual statistical 
models, namely on a real geophysical model \cite{Boore2003}.
We believe that the proposed method raises a range of interesting questions and directions of research;
we briefly discuss them in Section \ref{sec:discussion}.

\section{Proposed Scheme for Approximate Variational Inference}
\label{sec:proposed_scheme}

Assume an observed dataset of inputs $\bd{X}=(x_1,\dots,x_N)^T$ and outputs 
$\bd{Y}=(y_1,\dots,y_N)^T$ modelled by a model $f$ parametrised by $\bd{w}\in\RR{M}$. 
For observed outputs corrupted by Gaussian noise of precision $\beta$, the following 
likelihood\footnote{Besides the likelihood based on the Gaussian density, others can also be accommodated
as shown in Sections \ref{sec:logistic_regression}-\ref{sec:denoising}.
The choice of the Gaussian is made for ease of exposition.} arises:
\begin{eqnarray}
p(\bd{Y}|\bd{X},\bd{w},\beta) &=&  \prod_{n=1}^N{\mathcal N}(y_n|f(x_n;\bd{w}), \beta^{-1}) \notag\\
                    &=&  {\mathcal N}(\bd{Y}|f(\bd{X};\bd{w}), \beta^{-1} \bd{I}_N) \ ,
\end{eqnarray}
where $f(\bd{X};\bd{w})=(f(x_1;\bd{w}),\dots,f(x_N;\bd{w}))$ is the vector of model outputs calculated on all inputs $x_n$.
Furthermore, assume a Gaussian prior on the parameters $\bd{w}$:
\begin{equation}
p(\bd{w}| \alpha) = {\mathcal N}(\bd{w}|\bd{0},\alpha^{-1}\bd{I}_M) \ .
\end{equation}

Our wish is to approximate the true posterior of the parameters $p(\bd{w}|\bd{Y},\bd{X},\alpha,\beta)$.
We do not make any assumptions about the model having conjugate priors for the
parameters $\bd{w}$. Model $f$ may have a complex functional form that
hinders exact Bayesian inference, or even the application of an approximate
Bayesian scheme such as VBEM \cite{Beal2003} with a factorised prior.
However, we do have to make an assumption on the form of the posterior.
We choose to postulate an approximate Gaussian posterior for the parameters \bd{w}: 
\begin{equation}
q(\bd{w}) = {\mathcal N}(\bd{w}|\bd{\mu},\bd{\Sigma}) \ .
\end{equation}
Parameters $\bd{\mu}\in\RR{M}$ and $\bd{\Sigma}\in\RR{M\times M}$ of the posterior are called variational parameters. For reasons that will become obvious shortly, we choose to
parametrise covariance matrix as $\bd{\Sigma}=\bd{L}\bd{L}^T$
with $\bd{L}\in\RR{M\times M}$.
The postulated posterior now reads:
\begin{eqnarray}
q(\bd{w}) = {\mathcal N}(\bd{w}|\bd{\mu},\bd{L}\bd{L}^T) \ .
\end{eqnarray}
Hence, the actual variational parameters are $\bd{\mu}$ and $\bd{L}$.

\subsection{Approximate Lower Bound}

The first step in introducing the proposed scheme, is writing the marginal log-likelihood
and lower-bounding it in the standard way using Jensens' inequality \cite[Eq. (2.46)]{Beal2003}:

\begin{eqnarray}
\log p(\bd{Y}|\bd{X},\alpha, \beta)
&=&    \log \int p(\bd{Y}|\bd{X},\bd{w}, \beta) p(\bd{w}|\alpha)\bd{dw} \notag\\
&=&    \log \int \frac{q(\bd{w})}{q(\bd{w})} p(\bd{Y}|\bd{X},\bd{w}, \beta) p(\bd{w}|\alpha)\bd{dw} \notag\\
&\geq& \int q(\bd{w}) \log  \frac{p(\bd{Y}|\bd{X},\bd{w}, \beta) p(\bd{w}|\alpha)}{q(\bd{w})}\bd{dw} \notag\\
&=&    \int q(\bd{w}) \log  p(\bd{Y}|\bd{X},\bd{w}, \beta) \bd{dw} \notag\\
& & +  \int q(\bd{w}) \log \frac{p(\bd{w}|\alpha)}{q(\bd{w})}\bd{dw} \notag\\
&=&    \underbrace{ \int q(\bd{w}) \log  p(\bd{Y}|\bd{X},\bd{w}, \beta) \bd{dw} }_{(1)} \notag\\
& & -  \underbrace{ \int q(\bd{w}) \log \frac{q(\bd{w})}{p(\bd{w}|\alpha)}\bd{dw} }_{(2)} \notag \\
&\triangleq& \mathcal{L}(\bd{\mu},\bd{L},\alpha, \beta) \ .
\label{eq:lower_bounded_marg_likel}
\end{eqnarray}

An alternative motivation of the lower bound is provided in \cite[Eq. (15)]{Tzikas2008}
Maximising the lower bound $\mathcal{L}$ in 
Eq. (\ref{eq:lower_bounded_marg_likel}) with respect to
the free  variational parameters \bd{\mu} and \bd{L} of $q(\bd{w})$,
results in the best Gaussian approximation to the true posterior.
Term $(1)$, the integrated likelihood in Eq. (\ref{eq:lower_bounded_marg_likel}), is a potentially intractable integral.
We approximate term $(1)$ using Monte Carlo sampling:
\begin{eqnarray}
\frac{1}{S} \sum_{s=1}^S \log p(\bd{Y}|\bd{X}, \bd{w}_{(s)}, \beta) \ ,
\label{eq:sampled_term1}
\end{eqnarray}
where we draw $S$ samples $\bd{w}_{(s)}$ from the postulated posterior $q(\bd{w})$.
Due to the sampling, however, the variational parameters no longer appear in the approximation Eq. (\ref{eq:sampled_term1}) .
Nevertheless, it is possible to re-introduce them by rewriting the sampled weights $\bd{w}_{(s)}$ as\footnote{This is where the parametrisation $\bd{\Sigma}=\bd{L}\bd{L}^T$ becomes useful.}:
\begin{eqnarray}
\bd{w}_{(s)} = \bd{\mu} + \bd{L}\bd{z}_{(s)} \ ,
\end{eqnarray} 
where variables $\bd{z}_{(s)}$ are sampled from the standard normal 
$\bd{z} \sim \mathcal{N}(\bd{0},\bd{I}_M)$.
We summarise all samples $\bd{z}_{(s)}$ by $Z=\{\bd{z}_{(1)} \dots,\bd{z}_{(S)}\}$.
Hence, we can rewrite  Eq. (\ref{eq:sampled_term1}) as:
\begin{eqnarray}
\frac{1}{S} \sum_{s=1}^S \log p(\bd{Y}|\bd{X},\bd{\mu} + \bd{L}\bd{z}_{(s)},\beta) \ .
\label{eq:sampled_term1_explicit}
\end{eqnarray}
Hence, the variational parameters \bd{\mu} and \bd{L} are now made explicit in this approximation.
We expand the approximation of term $(1)$ further:
\begin{eqnarray}
& &\frac{1}{S} \sum_{s=1}^S \log p(\bd{Y}|\bd{X} ,\bd{\mu} + \bd{L}\bd{z}_{(s)}, \beta) = \notag \\ 
& & \frac{1}{S} \sum_{s=1}^S \log \mathcal{N}(\bd{Y}| f(\bd{X}; \bd{\mu} + \bd{L}\bd{z}_{(s)}),\beta^{-1}\bd{I}_N) = \notag \\
& & \frac{1}{S} \sum_{s=1}^S  \frac{N}{2}\log(\beta)  - \frac{\beta}{2}\|\bd{Y}-f(\bd{X}; \bd{\mu} + \bd{L}\bd{z}_{(s)}) \|^2  \notag \\
& & +\ const \ .
\label{eq:blr_approx_term1}
\end{eqnarray}

Term $(2)$ in Eq. (\ref{eq:lower_bounded_marg_likel}) is simply the Kullback-Leibler divergence (KLD) between the two Gaussian densities $q(\bd{w})$ and
$p(\bd{w}|\alpha)$, and can be calculated in closed form:
\begin{eqnarray}
\frac{1}{2}\bigg( tr(\alpha \bd{L}^T\bd{L})  + \alpha \bd{\mu}^T \bd{\mu} - M - \ln | \alpha\bd{L}\bd{L}^T | \bigg)\ .
\label{eq:kld_term}
\end{eqnarray}

We can now put together the approximated term $(1)$ in Eq. (\ref{eq:blr_approx_term1}) and the KLD term $(2)$ in
Eq.  (\ref{eq:kld_term}), to formulate the following objective function\footnote{The subscript $FS$ stands for finite sample.}
 ${\mathcal L}_{(FS)}$. Discarding constants,
the proposed approximate lower bound reads:
\begin{eqnarray}
& & {\mathcal L}_{(FS)}(\bd{\mu},\bd{L},\alpha, \beta, Z) =  \notag\\
& & \frac{1}{S} \sum_{s=1}^S \frac{N}{2}\log(\beta)  - \frac{\beta}{2}\|\bd{Y}-f(\bd{X}; \bd{\mu} + \bd{L}\bd{z}_{(s)}) \|^2 \notag\\
& & - \frac{1}{2}\bigg( tr(\alpha \bd{L}\bd{L}^T)  + \alpha \bd{\mu}^T \bd{\mu} - \ln | \alpha\bd{L}\bd{L}^T | \bigg)  \ .
\label{eq:approx_objective}
\end{eqnarray}
Objective ${\mathcal L}_{(FS)}$ is an approximation to the intractable lower bound ${\mathcal L}$ in 
Eq. (\ref{eq:lower_bounded_marg_likel}).
It consists of two parts, the approximation to the integrated likelihood (term $(1)$)
and the exact KLD (term $(2)$).
The proposed lower bound ${\mathcal L}_{(FS)}$ becomes more accurate 
when the number $S$ of samples $\bd{z}_{(s)}$ is large.

\subsection{Optimisation of Approximate Lower Bound}

Gradients of ${\mathcal L}_{(FS)}$ can be calculated with respect to the variational
parameters \bd{\mu} and \bd{L} in order to find the best
approximating posterior $q(\bd{w})$:

\begin{eqnarray}
& & \nabla_{\bd{\mu}} \mathcal{L}_{(FS)}(\bd{\mu},\bd{L},\alpha, \beta, Z) = \notag\\
& &  \frac{1}{S}   \sum_{s=1}^S \beta (\nabla_{\bd{w}} f)_{(s)}^T (\bd{Y} - f (\bd{X};\bd{\mu} + \bd{L}\bd{z}_{(s)}))  
 -   \alpha \bd{\mu}  \ ,
\label{eq:lower_bound_gradient_wrt_mu} 
\end{eqnarray}
\begin{eqnarray}
& & \nabla_{\bd{L}} \mathcal{L}_{(FS)}(\bd{\mu},\bd{L},\alpha, \beta, Z) =  \notag\\
& & \frac{1}{S}  \sum_{s=1}^S \beta (\nabla_{\bd{w}} f)_{(s)}^T (\bd{Y}-f (\bd{X};\bd{\mu} + \bd{L}\bd{z}_{(s)})) \bd{z}_{(s)}^T  \notag\\
& & -  \alpha\bd{L}  + {\bd{L}^+}^T \ ,
\label{eq:lower_bound_gradient_wrt_L} 
\end{eqnarray}
where $\nabla_{\bd{w}} f$ denotes the Jacobian matrix of $f$ and $\bd{L}^+$ is the pseudo-inverse of \bd{L}
due to the derivation of the 
log-determinant\footnote{See derivative rule $55$ in \cite{MatrixCookbook}.}
in Eq. (\ref{eq:approx_objective}).
Analogous equations for the case of exact variational inference can be found at \cite[Eq. (2.64)]{Beal2003}.
Given the current posterior $q(\bd{w})$, hyperparameters
$\alpha$ and $\beta$ have analytical updates:

\begin{equation}
\alpha = \frac{M}{\bd{\mu}^T \bd{\mu} + \mathrm{tr}(\bd{L}\bd{L}^T)} \ ,
\label{eq:approx_alpha_update}
\end{equation}

\begin{equation}
\beta = \frac{SN}{ \sum_{s=1}^S  \| \bd{Y}- f (\bd{X} ; \bd{\mu} + \bd{L}\bd{z}_{(s)}) \|^2  } \ .
\label{eq:approx_beta_update}
\end{equation}
Again, analogous equations for the above hyperparameter updates can be found in \cite[Eqs. (38), (39)]{Tzikas2008}.

The proposed scheme is summarised with the pseudocode in Algorithm \ref{proposed_scheme}.
Convergence was established in the experiments by checking whether the difference between the
objective function values ${\mathcal L}_{(FS)}$ between two successive iterations is less than $10^{-4}$.
Gradient optimisation of \bd{\mu} and \bd{L} was carried out using the scaled conjugate gradient algorithm \cite{Moller1993}.
The outcome of the above scheme is the approximation ${\mathcal L}_{(FS)}$ to the
marginal log-likelihood $\log p(\bd{Y}|\bd{X},\alpha,\beta)$ (also called log-evidence).
The scheme imparts us with the approximate Gaussian posterior $q(\bd{w})={\mathcal N}(\bd{w}|\bd{\mu},\bd{L}\bd{L}^T)$.


\begin{algorithm}[ht]
\caption{Proposed approximate variational inference}	
\label{proposed_scheme}
\begin{algorithmic}
\vspace{0.15cm}
\State {\bf Initialisation:}
\begin{itemize}
\item Initialise variational parameters e.g. $\bd{\mu}\sim\mathcal{N}(\bd{0},\bd{I})$ and $\bd{L} = c\bd{I}$ . \% e.g. $c=0.1$ \newline
Initialise hyperparameters e.g. $\alpha=\beta=0.1$. 
\item Alternative to above, ML estimates may be useful as initial values.
\item  Draw $S$ samples $\bd{z} \sim \mathcal{N}(0,\bd{I})$ that remain fixed throughout the algorithm.
\end{itemize}

\vspace{0.15cm}
\State {\bf Alternating optimisation of $\mathcal{L}_{(FS)}$:}

\For{$\mbox{iter}=1:\mbox{MaxIter}$ }  \% e.g. $MaxIter=1000$

	\begin{itemize}

		\item Record $L_{prv} \leftarrow \mathcal{L}_{(FS)}$. 
	
		\item Optimise \bd{\mu} for $J$ iterations  using the gradient
	  in Eq. (\ref{eq:lower_bound_gradient_wrt_mu}). \% e.g. $J=10$
	
		\item Optimise  \bd{L} for $J$ iterations  using the gradient
	  in Eq. (\ref{eq:lower_bound_gradient_wrt_L}).	 
	
		\item Update $\alpha$ and $\beta$ using Eqs. (\ref{eq:approx_alpha_update})
		and (\ref{eq:approx_beta_update}) respectively.
	
		\item Record $L_{new} \leftarrow \mathcal{L}_{(FS)}$. 
	
		\item Break if e.g. $L_{new} - L_{prv} <10^{-4}$. 

	\end{itemize}
	
\EndFor	

\vspace{0.15cm}
\State {\bf Result:}
\begin{itemize}
\item Lower bound $\mathcal{L}_{(FS)}$ to marginal log-likelihood.
\item Gaussian Posterior $\mathcal{N}(\bd{\mu},\bd{LL}^T)$.
\end{itemize}
\end{algorithmic}
\end{algorithm}

\subsection{Monitoring Generalisation Performance}
\label{sec:generalisation}

For large values of $S$  the proposed
lower bound $\mathcal{L}_{(FS)}$ approximates the true bound $\mathcal{L}$ in Eq. (\ref{eq:lower_bounded_marg_likel}) closely.
Therefore, we expect that optimising $\mathcal{L}_{(FS)}$ will yield
approximately the same optimal variational parameters $\bd{\mu}, \bd{L}$
as the optimisation of the intractable true lower bound $\mathcal{L}$ would.

The proposed scheme exhibits some fluctuation  
as $\mathcal{L}_{(FS)}(\bd{\mu},\bd{L},\alpha,\beta,Z)$ is a function 
of the random set of samples $Z$. Hence, if the algorithm, as summarised in Algorithm \ref{proposed_scheme},
is run again, a new set $Z^{(new)}$ will be drawn and a different 
function $\mathcal{L}_{(FS)}(\bd{\mu},\bd{L},\alpha,\beta,Z^{(new)})$ will be optimised.
However, for large enough $S$ the fluctuation due to $Z$ will be innocuous and approximately
the same variational parameters will be found for any drawn $Z$\footnote{Discounting other
sources of randomness like initialisation, etc.}.

However, if on the other hand we choose a small $S$, then the variational parameters
will overly depend on the small set of samples $Z$ that happened to be drawn at the beginning of the algorithm.
As a consequence, $\mathcal{L}_{(FS)}$ will
approximate $\mathcal{L}$ poorly, and the resulting posterior $q(\bd{w})$ will also be a poor approximation to the true posterior.
Hence, the variational parameters will be overfitted
to the small set of samples $Z$ that happened to be drawn. 

Naturally, the question arises of how to choose a large enough $S$ in order avoid overfitting the
variational parameters on $Z$.
A practical answer to this question is the following: at the beginning of the algorithm we draw 
a second independent set of samples $Z^{\prime}=\{\bd{z}^{\prime}_{(1)}, \dots,\bd{z}^{\prime}_{(S^{\prime})}\}$
where $S^{\prime}$ is preferably a number larger than $S$. At each (or every few) iteration(s) 
we monitor the quantity $\mathcal{L}_{(FS)}(\bd{\mu},\bd{L},\alpha,\beta,Z^{\prime})$ on the
independent\footnote{We stress that $Z^{\prime}$ is not used in training.} sample set $Z^{\prime}$. 
If the variational parameters are {\itshape not overfitting} the drawn $Z$, then 
we should see that as the lower bound $\mathcal{L}_{(FS)}(\bd{\mu},\bd{L},\alpha,\beta,Z)$
increases, the quantity $\mathcal{L}_{(FS)}(\bd{\mu},\bd{L},\alpha,\beta,Z^{\prime})$ should also display a tendency to increase. If on the other hand the variational parameters {\itshape are overfitting} the drawn $Z$,
then though $\mathcal{L}_{(FS)}(\bd{\mu},\bd{L},\alpha,\beta,Z)$
is increasing, we will notice that $\mathcal{L}_{(FS)}(\bd{\mu},\bd{L},\alpha,\beta,Z^{\prime})$ 
is actually deteriorating. This is a clear sign that  a larger $S$ is required.

The described procedure is reminiscent of monitoring the generalisation performance of a learning
algorithm on a validation set during training. A significant difference, however, is that while
validation sets are typically of limited size, here we can set $S^{\prime}$ arbitrarily large.
For practical purposes, we found that $S^{\prime}=5S$ was good enough to detect overfitting.
An illustration of overfitting the variational parameters in provided in Sec. \ref{sec:blr}.

\section{Applications}
\label{sec:applications}

In this section we apply the proposed approach on a variety of applications,
namely regression, classification, denoising and geophysical modelling.
In particular, the geophysical example shows how the method can be applied 
beyond standard statistical models.

\subsection{Fitting Bivariate Posteriors}
\label{sec:fitting_bivariate}

\begin{figure}[!t]
\centering
\subfloat[Proposed, KLD=0.351\ .]{%
\includegraphics[scale=0.3]{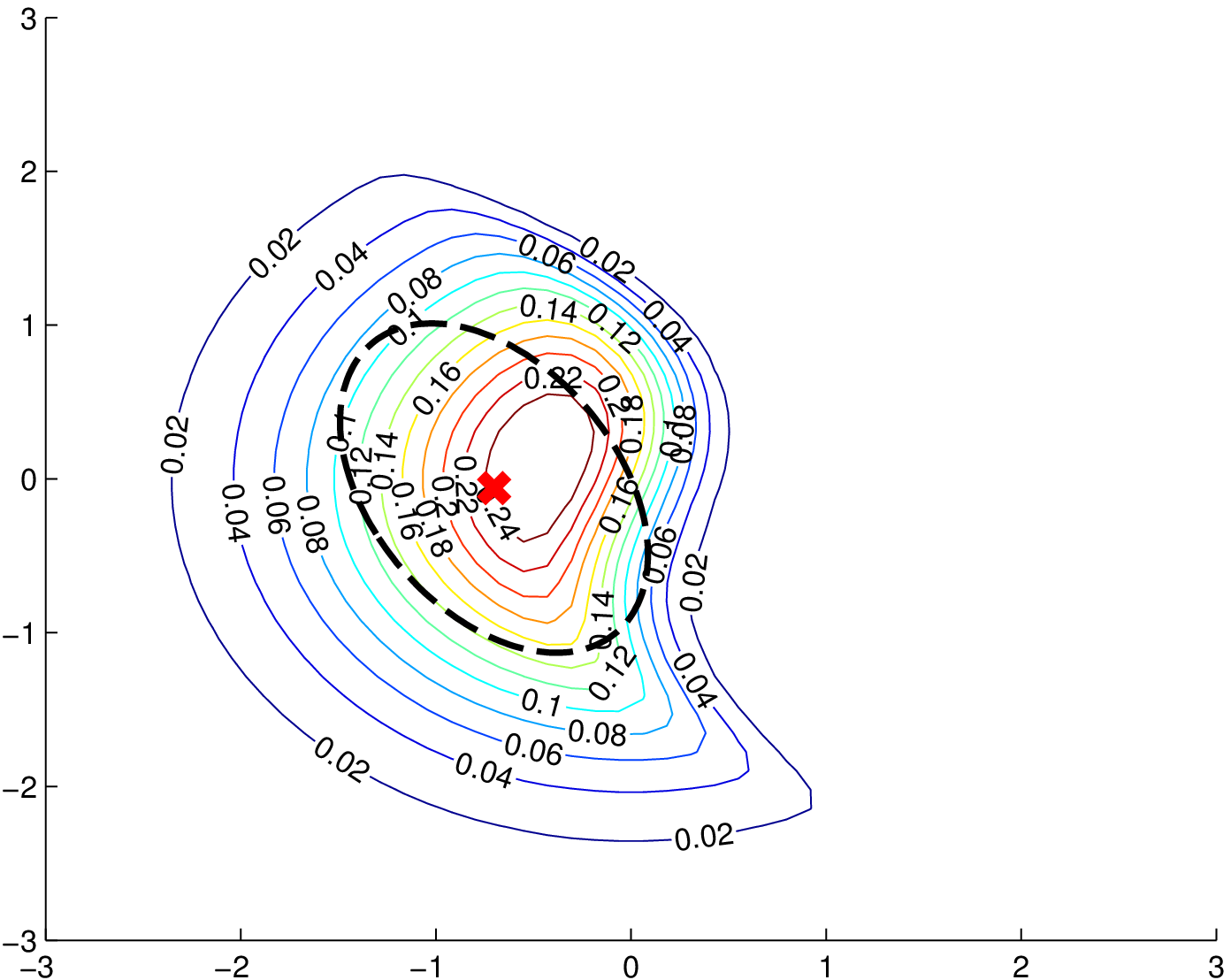}
}
\subfloat[Laplace, KLD=4.570\ .]{%
\includegraphics[scale=0.3]{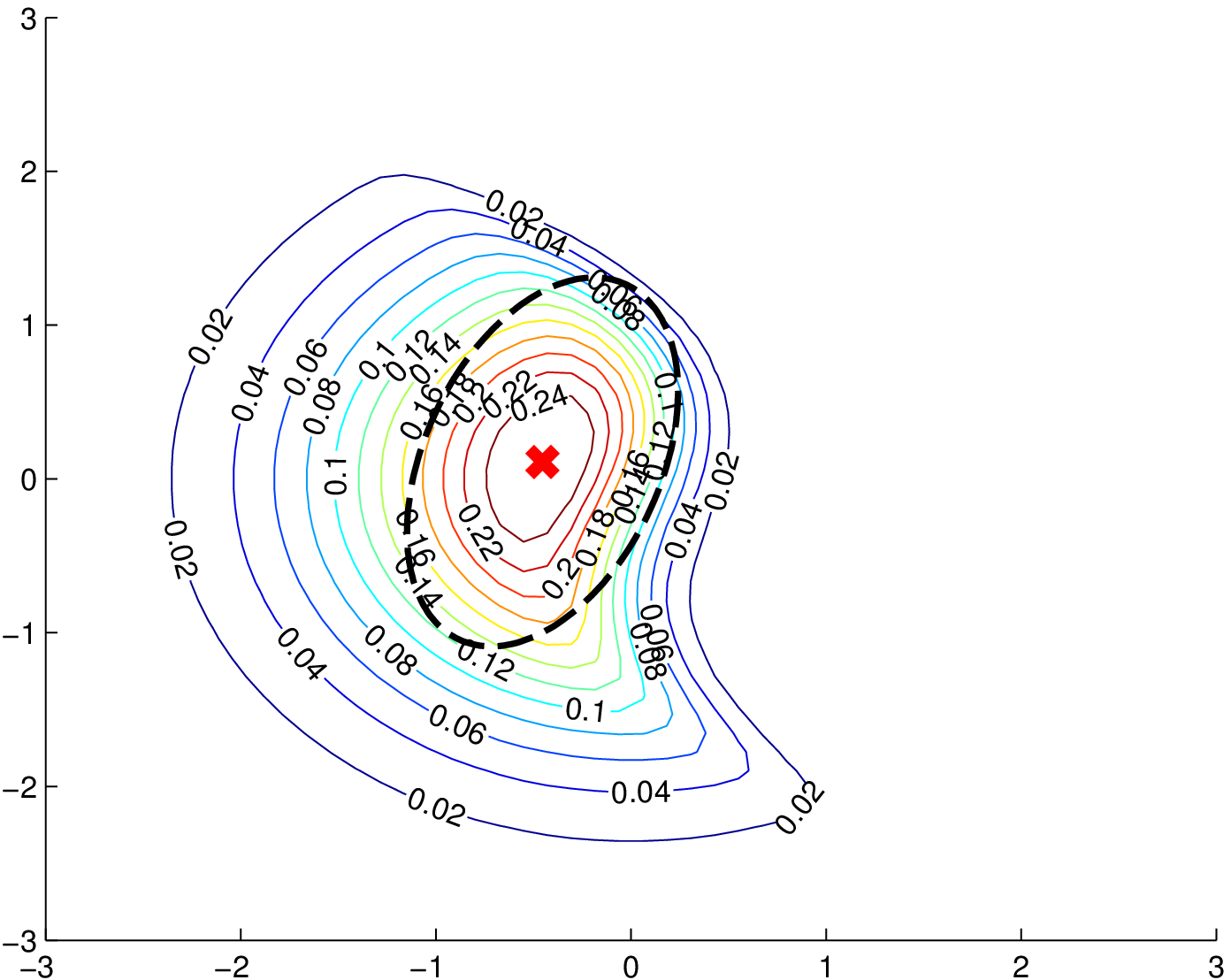}
} \\
\subfloat[Proposed, KLD=0.585\ .]{%
\includegraphics[scale=0.3]{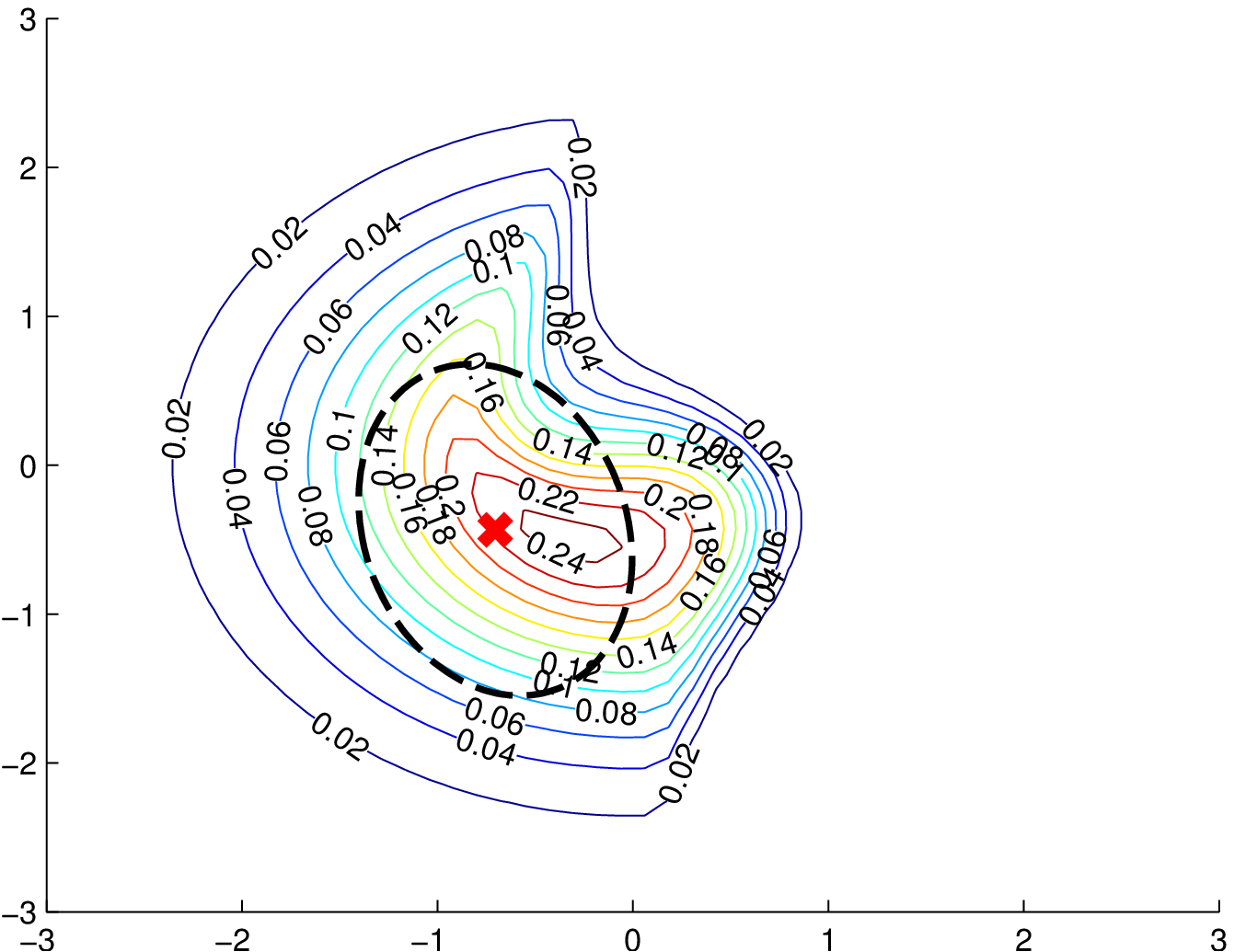}
}
\subfloat[Laplace, KLD=13.915\ .]{%
\includegraphics[scale=0.3]{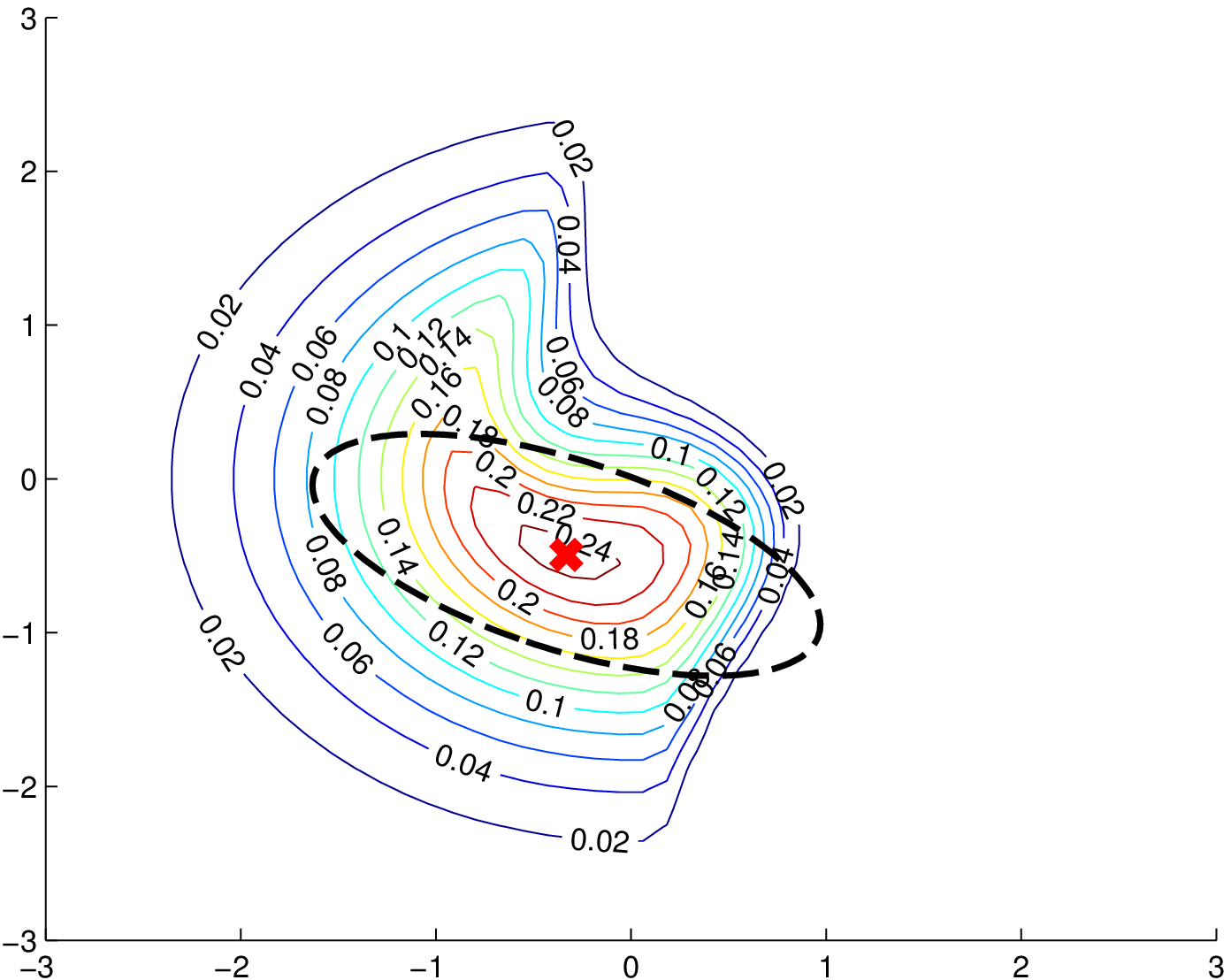}
} \\
\subfloat[Proposed, KLD=1.103\ .]{%
\includegraphics[scale=0.3]{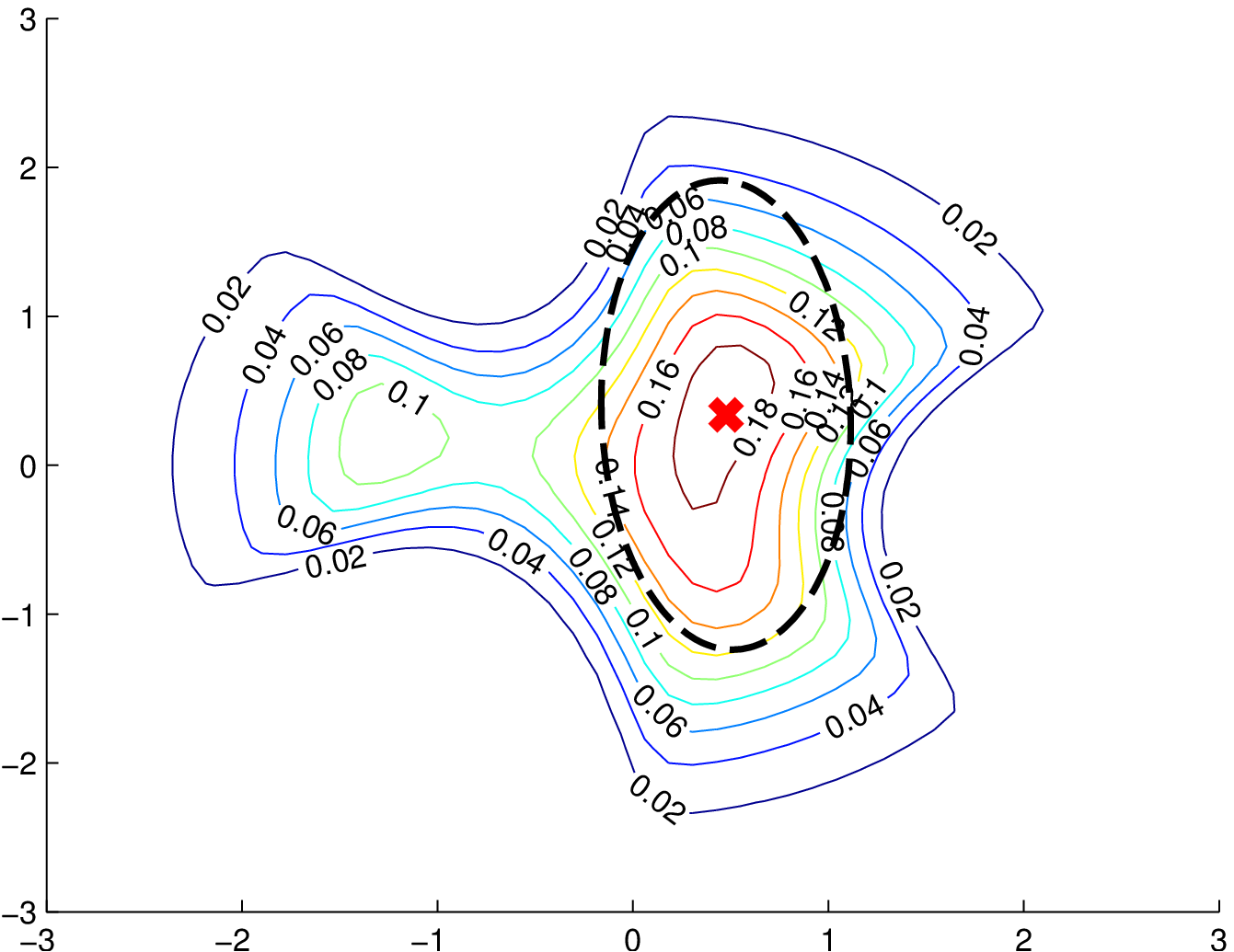}
}
\subfloat[Laplace, KLD=1.384\ .]{%
\includegraphics[scale=0.3]{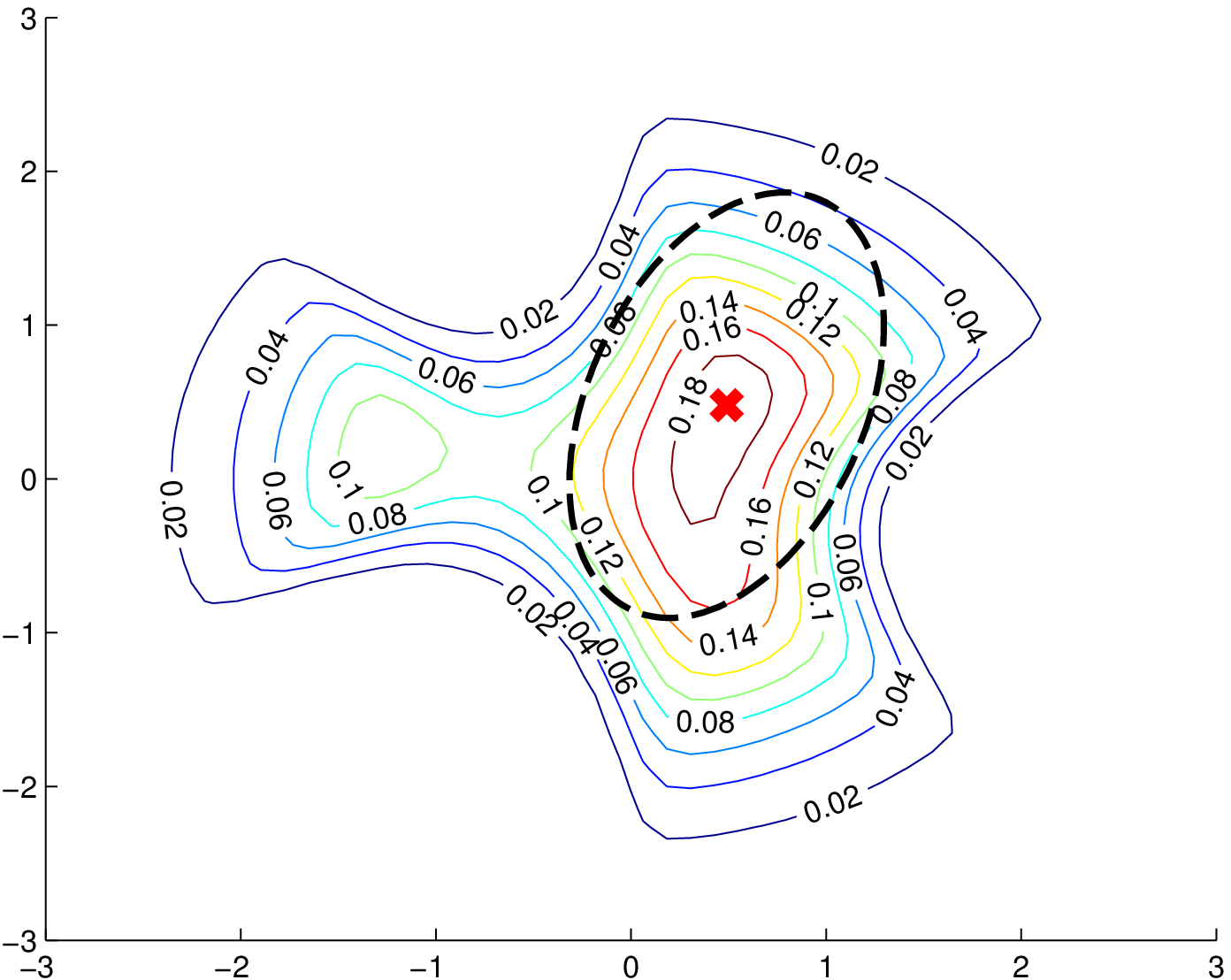}
}
\caption{Top  $\bd{a}=(-3,1,-1,-1,-1,-1)^T$, middle  $\bd{a}=(0,-2,-4,-1,-3,0)^T$, bottom  $\bd{a}=(1,0,2,1,-1,0)^T$.
The KLD values reveal (lower is better) that the proposed scheme fares better than the Laplace approximation.}
\label{fig:approximations_bivariate}
\end{figure}

We test our proposed scheme on some artificially constructed posteriors
by using a flexible parametric class of densities, due to \cite{Azzalini2005},
which reads:
\begin{equation}
f(w_1,w_2) = 2{\mathcal N}(w_1,w_2|0,\bd{I}_2) \Phi (h(w_1,w_2)) \ ,
\label{eq:flexible_class}
\end{equation}
where $\Phi$ is the cumulative distribution function
of the standard normal distribution, and in general
$h$ is a real-value function such that $h(-w)=-h(w)$.
Here, we take $h$ to be the dot product of a row and column vector, $h(w_1,w_2)=(w_1,w_2,w_1 w_2^2,w_1^2 w_2,w_1^3,w_2^3)\bd{a}$ as in \cite{Azzalini2005}.
The goal is to find the best Gaussian approximation to
instances of Eq. (\ref{eq:flexible_class}) for different column vectors
\bd{a}. To that end, we tried to find the best Gaussian using
the Laplace approximation and the proposed scheme. We used $S=50$.
The results are shown in Fig. \ref{fig:approximations_bivariate}.
The Gaussian approximations are drawn as  black dashed curves with their
mean marked as a red cross.
The goodness of each approximation has been measured as the Kullback-Leibler
divergence, and it is noted in the respective captions.
We note that the proposed scheme fares better than the Laplace
approximation as the latter evidently focuses on the mode
of the target density instead on where the volume of the density lies.
The KLD in these examples was
calculated numerically as there is no closed form
between a Gaussian and a member of the densities in 
Eq. (\ref{eq:flexible_class})

\subsection{Bayesian Linear Regression}
\label{sec:blr}

\begin{figure*}[t]
\centering
\includegraphics[width=0.5\textwidth]{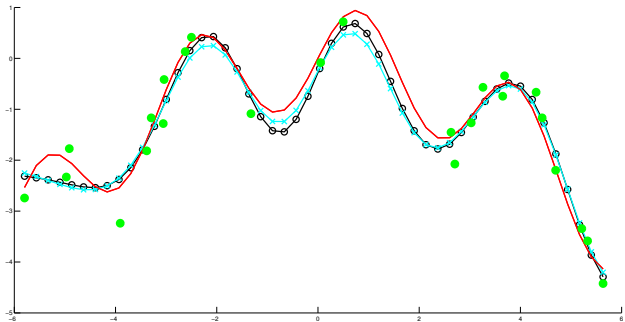}
\caption{Mean predictions.
The red line is the true underlying function, and the green points are
noisy realisations of it (data items). 
The black (circles) and cyan (crosses) lines are the mean predictions that correspond
to exact and approximate inference respectively. We see that the approximate scheme
stands in close agreement to the exact solution.}
\label{fig:blr_solutions}
\end{figure*}

\begin{figure*}
\centering
\subfloat[Exact inference.]{%
\includegraphics[width=0.25\textwidth]{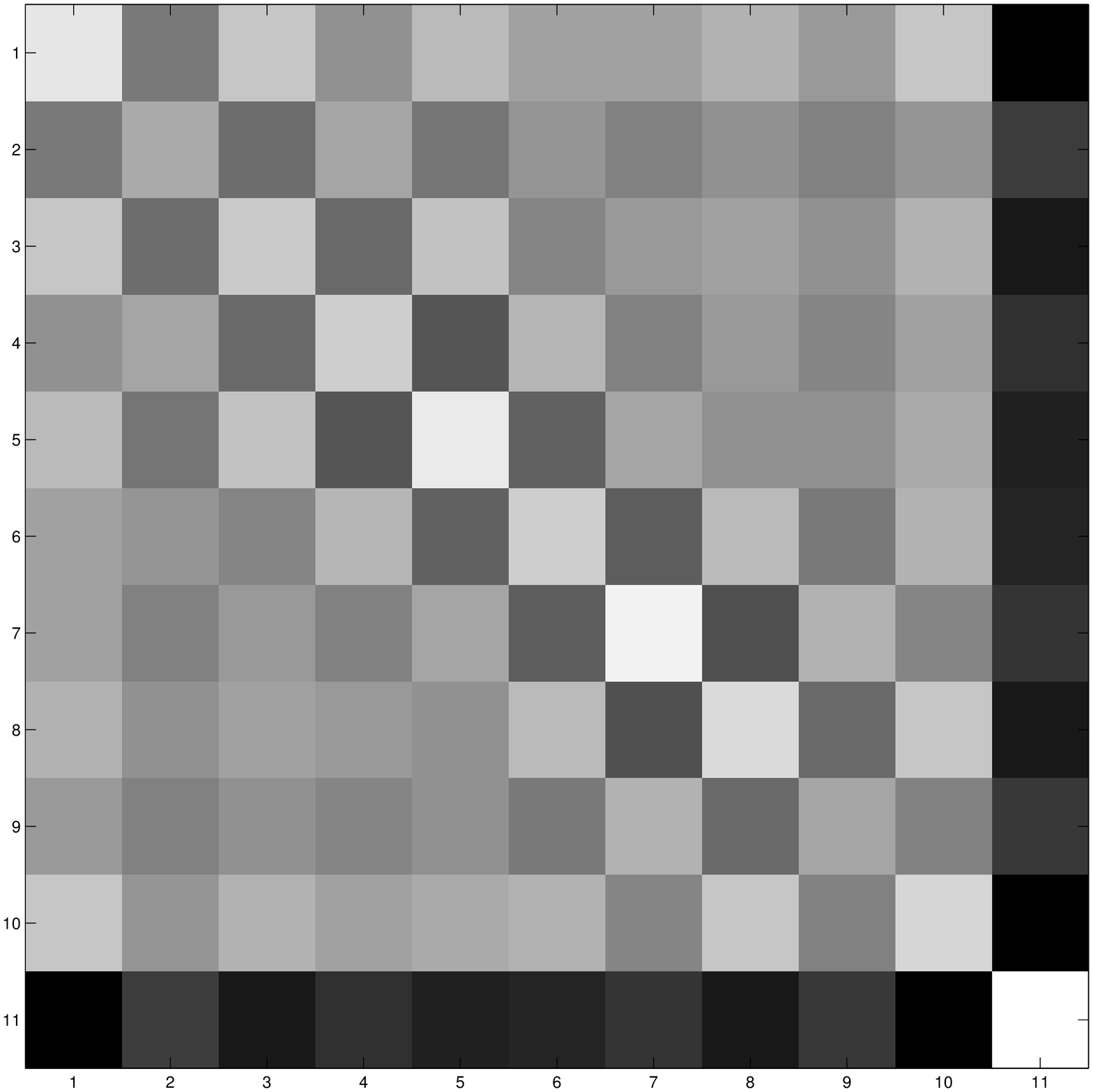}
}
\subfloat[Proposed.]{%
\includegraphics[width=0.25\textwidth]{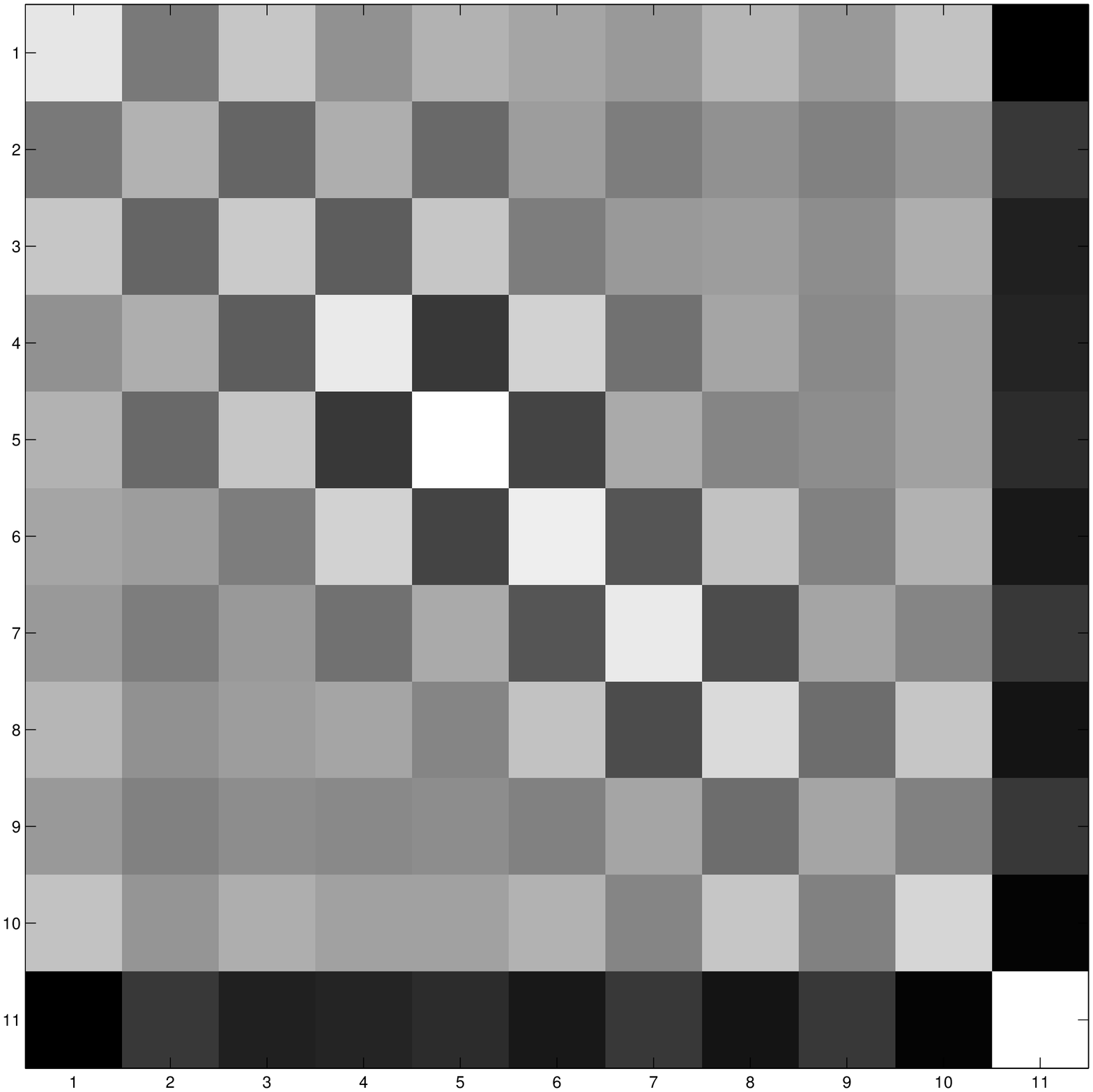}
}
%
\caption{Posterior covariance matrices found by both schemes. We note their close similarity
which indicates that the proposed scheme stands in close agreement to the exact solution.}
\label{fig:blr_cov_matrices}
\end{figure*}

Bayesian linear regression constitutes a useful example 
for corroborating that the proposed scheme works correctly as
we can compare the obtained posterior $q(\bd{w})$ 
to the posterior obtained by \emph{exact} Bayesian inference \cite{Bishop}.

We consider data targets  $y_n$
generated by the equation 
\begin{equation}
y = 2\cos(x)\sin(x)-0.1x^2 \ ,
\end{equation}
with inputs $x_n$ uniformly drawn in the range $[-6,6]$.
We add white Gaussian noise to the data targets 
with a standard deviation of $\sigma = 0.2$.
We calculate a set of radial basis functions on the data inputs
\[ \bd{\phi}_n = [ \phi(x_n;r,c_1) \dots \phi(x_n;r,c_{M-1}) \ 1]^T \]
where $\phi(x_n;r,c_m) = \exp(-\frac{\|x_n-c_m\|^2}{2r^2})$.
The last element $1$ in $\bd{\phi}_n$  serves as a  bias term.
We set $r=1$, and adopt the linear model $y=\bd{\phi}^T\bd{w}$
where $\bd{w}\in\RR{M}$.
We complete the model by choosing the following densities:

\begin{figure*}
\centering
\subfloat[S=100, no overfitting.]{%
\includegraphics[width=0.40\textwidth]{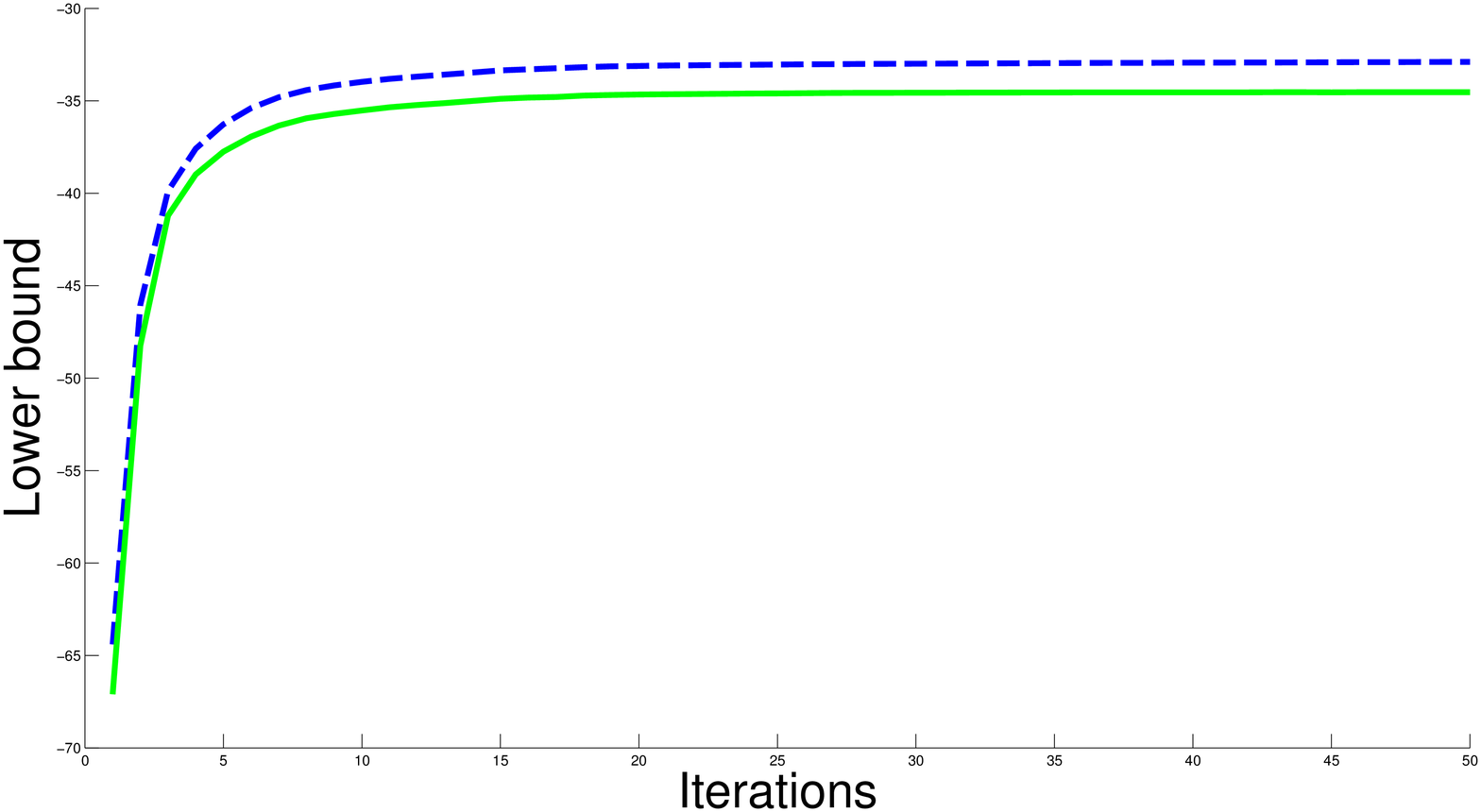}
}
\subfloat[S=10, overfitting.]{%
\includegraphics[width=0.40\textwidth]{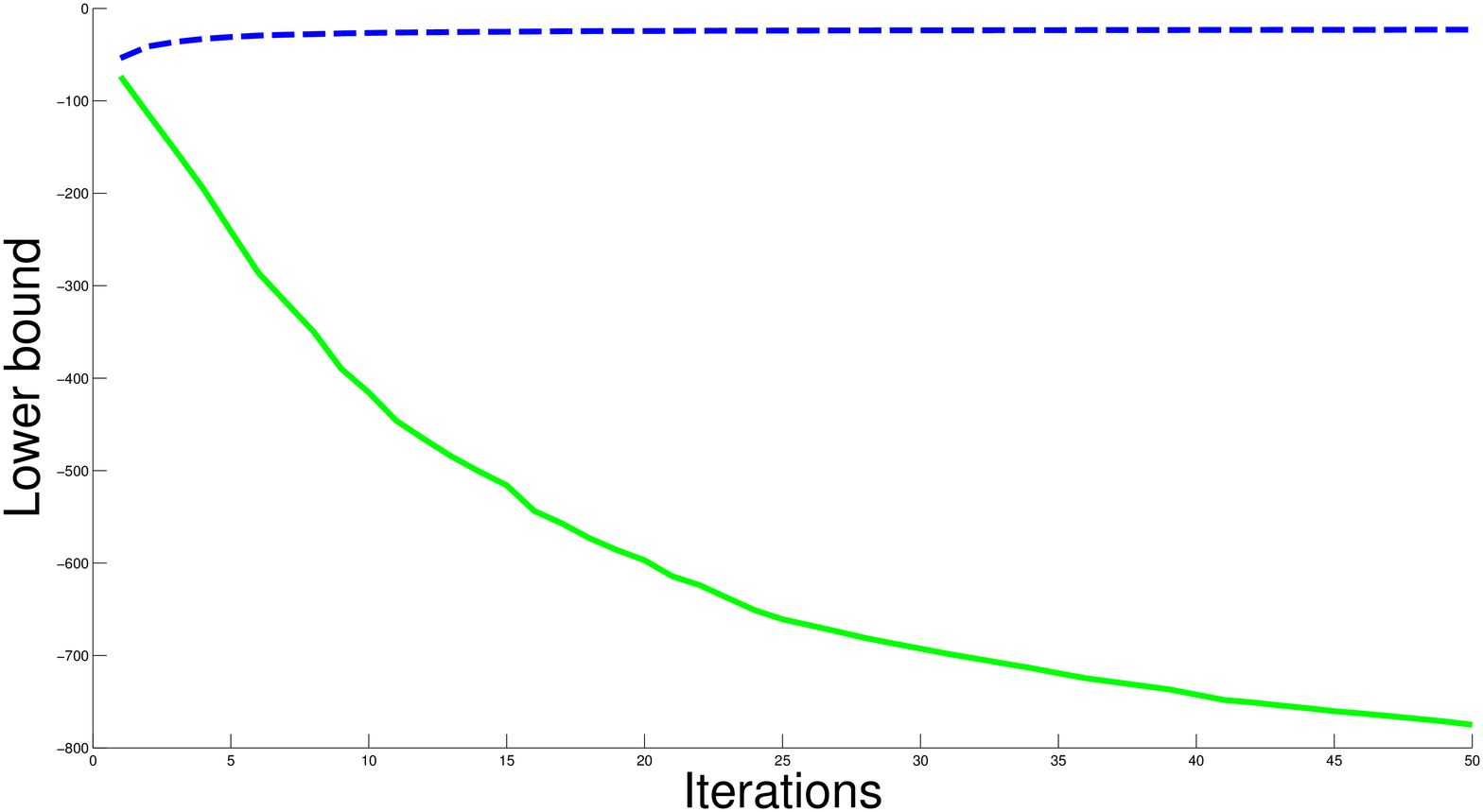}
}
%
\caption{Monitoring generalisation performance. Green solid line is $\mathcal{L}_{(FS)}(\bd{\mu},\bd{L},\alpha,\beta,Z^{\prime})$, blue dashed line is $\mathcal{L}_{(FS)}(\bd{\mu},\bd{L},\alpha,\beta,Z)$.
Overfitting of the variational parameters occurs when the number $S$ of samples $\bd{z}$ is not large enough.}
\label{fig:blr_generalisation}
\end{figure*}


\begin{itemize}
\item Likelihood: 
$p(\bd{Y}|\bd{X}, \bd{w}, \beta) = \prod_{n=1}^N \mathcal{N}(y_n | \bd{\phi}_n^T \bd{w},\beta^{-1})$.
\item Prior: $p(\bd{w}) = \mathcal{N}(\bd{w} | \bd{0},\alpha^{-1}\bd{I}_M)$\ .
\item Postulated posterior: $q(\bd{w}) = \mathcal{N}(\bd{w} | \bd{\mu},\bd{L}\bd{L}^T)$, 
where $\bd{\mu}\in\RR{M}$ and $\bd{L}\in \RR{M\times M}$.
\end{itemize}

We set the number of samples of variables $\bd{z}$ to $S=100$.
We inferred the Gaussian posterior of the weights \bd{w} using both exact Bayesian inference  \cite{Bishop} and the proposed scheme. 
In Fig. \ref{fig:blr_solutions} we plot the mean predictions as obtained by the two schemes
and note that they are very similar, especially in the areas where enough data items are present.
Similarly, in Fig. \ref{fig:blr_cov_matrices} we plot the covariance matrices found by the two schemes
and note their close similarity. Hence, we conclude that the proposed  scheme  stands in close agreement
to the exact Bayesian solution.

Finally, we demonstrate on the same dataset the effect of overfitting  the variational parameters when $S$
is set too low. In Fig. \ref{fig:blr_generalisation}
we are monitoring the lower bounds $\mathcal{L}_{(FS)}(\bd{\mu},\bd{L},\alpha,\beta,Z)$ and $\mathcal{L}_{(FS)}(\bd{\mu},\bd{L},\alpha,\beta,Z^{\prime})$, see Sec. \ref{sec:generalisation}.
In the left, we run the algorithm for $S=100$ and $S^{\prime}=500$: we see that as $\mathcal{L}_{(FS)}(\bd{\mu},\bd{L},\alpha,\beta,Z)$ increases with each iteration, so does $\mathcal{L}_{(FS)}(\bd{\mu},\bd{L},\alpha,\beta,Z^{\prime})$.
This means that the fitted variational parameters $\bd{\mu}, \bd{L}$ generalise well.
On the right hand side, we run the algorithm for $S=10$ but kept $S^{\prime}=500$.
Here we clearly see that while $\mathcal{L}_{(FS)}(\bd{\mu},\bd{L},\alpha,\beta,Z)$ increases, 
the lower bound $\mathcal{L}_{(FS)}(\bd{\mu},\bd{L},\alpha,\beta,Z^{\prime})$ is 
deteriorating. This a clear sign that a larger $S$ is required and that the
variational parameters are overfitted.

\subsection{Bayesian Logistic Regression}
\label{sec:logistic_regression}

In this section we apply the proposed scheme to Bayesian logistic
regression and compare with the variational approach presented
in \cite{Jaakkola2000}. 
The data are input-label pairs $(\bd{x},y)$ with $y\in\{0,1\}$.
Again, like in Sec. \ref{sec:blr}, we calculate basis functions
$\bd{\phi}_n$ on the input data $\bd{x}_n$ and take $r=0.5$.
We set $S=200$.
We complete the model by choosing the following densities:

\begin{itemize}
\item $\begin{fleqn}[0pt]\begin{aligned}[t] \mbox{Likelihood:\ } & p(\bd{Y}|\bd{X}, \bd{w}) = \\ &\prod_{n=1}^N \sigma(\bd{\phi}_n^T \bd{w})^{y_n}  (1-\sigma(\bd{\phi}_n^T \bd{w}))^{1-y_n} \ . \end{aligned}\end{fleqn}$
%
\item Prior: $p(\bd{w}) = \mathcal{N}(\bd{w} | \bd{0},\alpha^{-1}\bd{I}_M)$\ .
\item Postulated posterior: $q(\bd{w}) = \mathcal{N}(\bd{w} | \bd{\mu},\bd{L}\bd{L}^T)$\ .
\end{itemize}

We evaluated both schemes on  datasets 
 preprocessed by R\"atsch et al\footnote{\url{http://www.raetschlab.org/Members/raetsch/benchmark}}.
Each preprocessed dataset has been randomly partitioned into a 100 non-overlapping
training and testing sets. Hence the performance of both schemes
was evaluated as the accuracy on the test set, that is
the ratio of correctly classified test samples over all test samples.
The predictive distribution for the proposed scheme was approximated
using a Monte Carlo estimate. We drew 200 parameter samples
from the fitted Gaussian posterior $q$ and
measured performance on the testing set as the average accuracy under each sample of parameters.
The results reported in Table \ref{tbl:results_logistic}, 
mean squared error and standard deviation, show that both the proposed
schemes and the variational bound in \cite{Jaakkola2000} perform virtually the same.
We note that, as opposed to \cite{Jaakkola2000} which exploits the functional form of logistic regression in order 
to design a bespoke lower bound, the proposed method does not take into account any such knowledge and still
is capable of delivering comparable performance. Hence, we find the results in this section encouraging.

\begin{table}[!t]
\caption{Classification performance for Bayesian logistic regression (higher is better).}
  \label{tbl:results_logistic}
\centering
  \begin{tabular}{| c | c | c |}
    \hline
    Dataset & Bound in \cite{Jaakkola2000} & Proposed \\
    \hline        
    Banana & 0.8893 $\pm$ 0.0055 &  0.8893 $\pm$ 0.0054  \\ \hline
    Cancer & 0.7119 $\pm$ 0.0456 &  0.7116 $\pm$ 0.0454  \\ \hline
    Heart  & 0.5528 $\pm$ 0.0445 &  0.5500 $\pm$ 0.0476  \\ \hline
    Solar  & 0.6449 $\pm$ 0.0172 &  0.6445 $\pm$ 0.0167  \\ \hline
  \end{tabular}
\end{table}

\subsection{Bayesian Multiclass Classification}
\label{sec:multiclass}

In this section we apply the proposed scheme on Bayesian multiclass classification.
The data are input-label pairs $(\bd{x},\bd{y})$.
Vectors $\bd{y}$ are binary vectors encoding class labels using $1$-of-$K$ coding scheme, e.g. $[0\ 1\ 0]$ encodes class label $2$ in a $3$-class problem.
A typical way of formulating multiclass classification is multiclass logistic regression (MLR), see \cite[Chapter $4$]{Bishop} for more details. MLR models the probability $p(C_k|\bd{\phi}_n)$ of the $n$-th data item belonging to class $C_k$ via the softmax function
 $p(C_k|\bd{\phi}_n)=\frac{\exp(\bd{\phi}_n^T\bd{w}_k)}{\sum_{\ell=1}^K \exp(\bd{\phi}_n^T\bd{w}_\ell)}$. $K$ denotes the total number of classes, and each class $C_k$ is associated with a weight vector $\bd{w}_k$.
Similarly to logistic regression, MLR does not allow direct Bayesian inference as the use of the softmax function renders integrals over the likelihood term intractable. Thus, Bayesian MLR is a good candidate problem for the proposed approach. We specify the following model:
\begin{itemize}
\item $\begin{fleqn}[0pt]\begin{aligned}[t] \mbox{Likelihood:\ } & p(\bd{Y}|\bd{X}, \bd{w}_1,\dots,\bd{w}_K) = \\ & \prod_{n=1}^N \prod_{k=1}^K p(C_k|\bd{\phi}_n)^{y_{nk}} \ . \end{aligned}\end{fleqn}$
%
\item Prior: $\prod_{k=1}^K p(\bd{w}_k) = \mathcal{N}(\bd{w}_k | \bd{0},\alpha^{-1}\bd{I}_M)$\ .
\item Postulated posterior: 
$q(\bd{w}_1,\dots,\bd{w}_K) = \prod_{k=1}^K q(\bd{w}_k)$,
with $q(\bd{w}_k) = \mathcal{N}(\bd{w}_k | \bd{\mu}_k,\bd{L}_k\bd{L}_k^T)$\ .
\end{itemize}

\begin{table*}[!t]
\caption{Classification performance for Bayesian multiclass classification (higher is better).}
  \label{tbl:results_multiclass}
\centering
  \begin{tabular}{| c | c | c | c | c | c |}
    \hline
    Dataset & C  & Suggested Kernel & $\mbox{mRVM}_2$  & Proposed \\
    \hline        
    Ecoli 		  &8&Gaussian	&$0.855 \pm 0.047$  &  $0.852\pm 0.060$   \\ \hline
    Glass 		  &6&Polynomial	&$0.581 \pm 0.143$  &  $0.667\pm 0.112$   \\ \hline
    Iris 		  &3&Gaussian	&$0.913 \pm 0.071$  &  $0.947\pm 0.053$   \\ \hline
    Wine			  &3&Linear		&$0.959 \pm 0.062$  &  $0.976\pm 0.030$   \\ \hline
    Soybean (small)&4&Linear		&$1.000 \pm 0.000$  &  $1.000\pm 0.000$   \\ \hline
    	Vehicle		  &4&Polynomial	&$0.481 \pm 0.054$  &  $0.539\pm 0.103$   \\ \hline
    	Balance		  &3&Polynomial	&$0.931 \pm 0.044$  &  $0.947\pm 0.029$   \\ \hline
    	Crabs		  &4&Linear		&$0.915 \pm 0.071$  &  $0.950\pm 0.033$   \\ \hline
  \end{tabular}
\end{table*}

As a corroboration of the usefulness of our approximation to Bayesian MLR, we compare with the multiclass
generalisation of the relevance vector machine (RVM) \cite{Tipping2001}  presented in \cite{Psorakis2010}. We use the multiclass
UCI datasets suggested therein.  Amongst the two generalisations of the RVM suggested in \cite{Psorakis2010}, we use the $\mbox{mRVM}_2$ version. We also follow the suggestion of \cite{Psorakis2010} concerning the choice of kernels for the different datasets. We set $S=200$. Table \ref{tbl:results_multiclass} summarises the results of our numerical simulations along with details of the datasets. While the $\mbox{mRVM}_2$ designs a refined probabilistic model in order to make probabilistic multiclass classification possible, the proposed scheme does not take into account any kind of such knowledge and is still able to  deliver competitive performance, in terms of predictive accuracy, as seen in Table \ref{tbl:results_multiclass}. The good performance demonstrates both the usefulness and versatility of the proposed method.

\subsection{Probabilistic Image Denoising}
\label{sec:denoising}

In this section we further demonstrate how the proposed method can take in its stride a change in the model-likelihood
that complicates computations. The model considered here is the ubiquitous probabilistic principal component analysis (PPCA) introduced
in \cite{Tipping1999}. 
PPCA assumes that the observed high-dimensional data $\bd{y}\in\RR{d}$ are manifestations of low-dimensional latent variables $\bd{x}\in\RR{q}$, under a linear mapping expressed by a matrix $\bd{W}\in\RR{d\times q}$ and an offset $\bd{\xi}\in\RR{d}$, corrupted by Gaussian noise \bd{\epsilon}:
\begin{equation}
\bd{y} = \bd{W}\bd{x} + \bd{\xi} + \bd{\epsilon} \ .
\label{eq:ppca_model}
\end{equation}

PPCA formulates a computationally amenable linear-Gaussian model which allows integrating out the latent variables \bd{x} and obtaining the marginal log-likelihood. Estimating \bd{W} and \bd{\mu} follows 
by maximising the marginal log-likelihood \cite{Tipping1999}. Various works extend PPCA by replacing the noise model  with other choices, e.g. \cite{Archambeau2006} uses the Student-t distribution, in order to deal with different types of noise. A recent interesting suggestion is the choice of the Cauchy density as the noise model \cite{Xie2015}, \emph{albeit in a  non-probabilistic formulation}.  The Cauchy density with location $x_0$ and scale $\gamma>0$ parameters reads:
\begin{equation}
\left(\pi\gamma \left[1 + \left(\frac{x-x_0}{\gamma}\right)^2 \right] \right)^{-1} \ .
\label{eq:cauchy}
\end{equation}

Choosing the Cauchy density as the noise model leads to a version of PPCA where the marginal log-likelihood is no longer tractable and so the latent variables \bd{x} cannot be integrated out. This is simply because the prior
on  \bd{x} is not conjugate to the Cauchy likelihood.
However, the proposed method can be used to approximate this intractable marginal log-likelihood.
Formally, we specify the following Cauchy-PPCA model:

\begin{itemize}
\item $\begin{fleqn}[0pt]\begin{aligned}[t] &\mbox{Likelihood:\ }  p(\bd{Y}|\bd{X}, \bd{W},\bd{\xi},\gamma) = \\
&\prod_{n=1}^N  \left(\pi\gamma \left[1 + \left(\frac{\bd{y}_n-\bd{W}\bd{x}_n - \bd{\xi}}{\gamma}\right)^2 \right] \right)^{-1} \ . \end{aligned}\end{fleqn}$
\item Prior: $p(\bd{X}) = \mathcal{N}(\bd{X}| \bd{0}, \bd{I}_N)$\ .
\item Postulated posterior: $q(\bd{X}) = \mathcal{N}(\bd{X} | \bd{\mu},\bd{L}\bd{L}^T)$\ .
\end{itemize}
Parameters $\bd{W}$, $\bd{\xi}$ and $\gamma$ are obtained by gradient optimisation of the proposed lower bound $\mathcal{L}_{(FS)}$.

\begin{figure*}[!t]
\captionsetup[subfigure]{labelformat=empty}
\centering
\subfloat[$0$]{%
\includegraphics[width=0.175\textwidth]{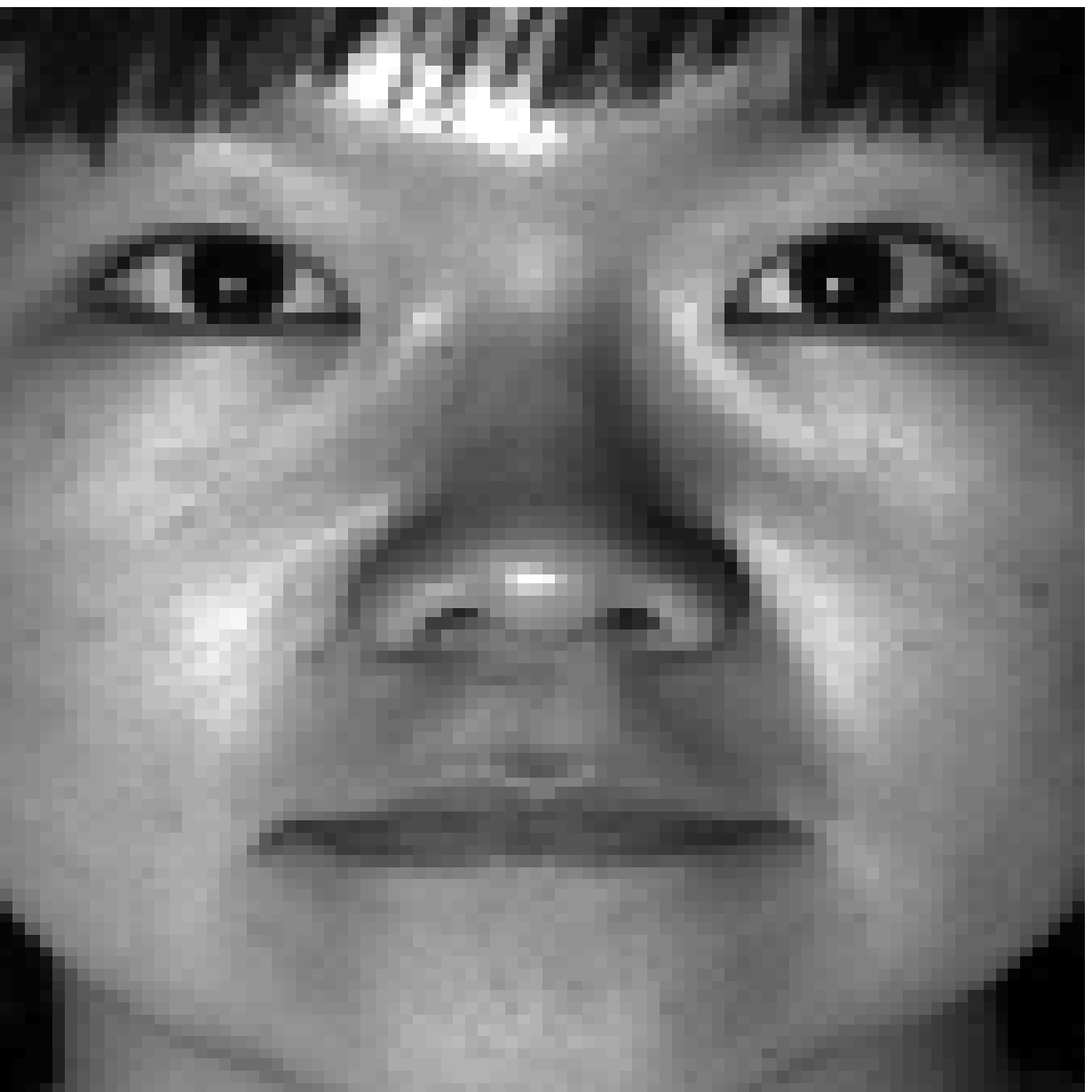}
}
\subfloat[$0.8999$]{%
\includegraphics[width=0.175\textwidth]{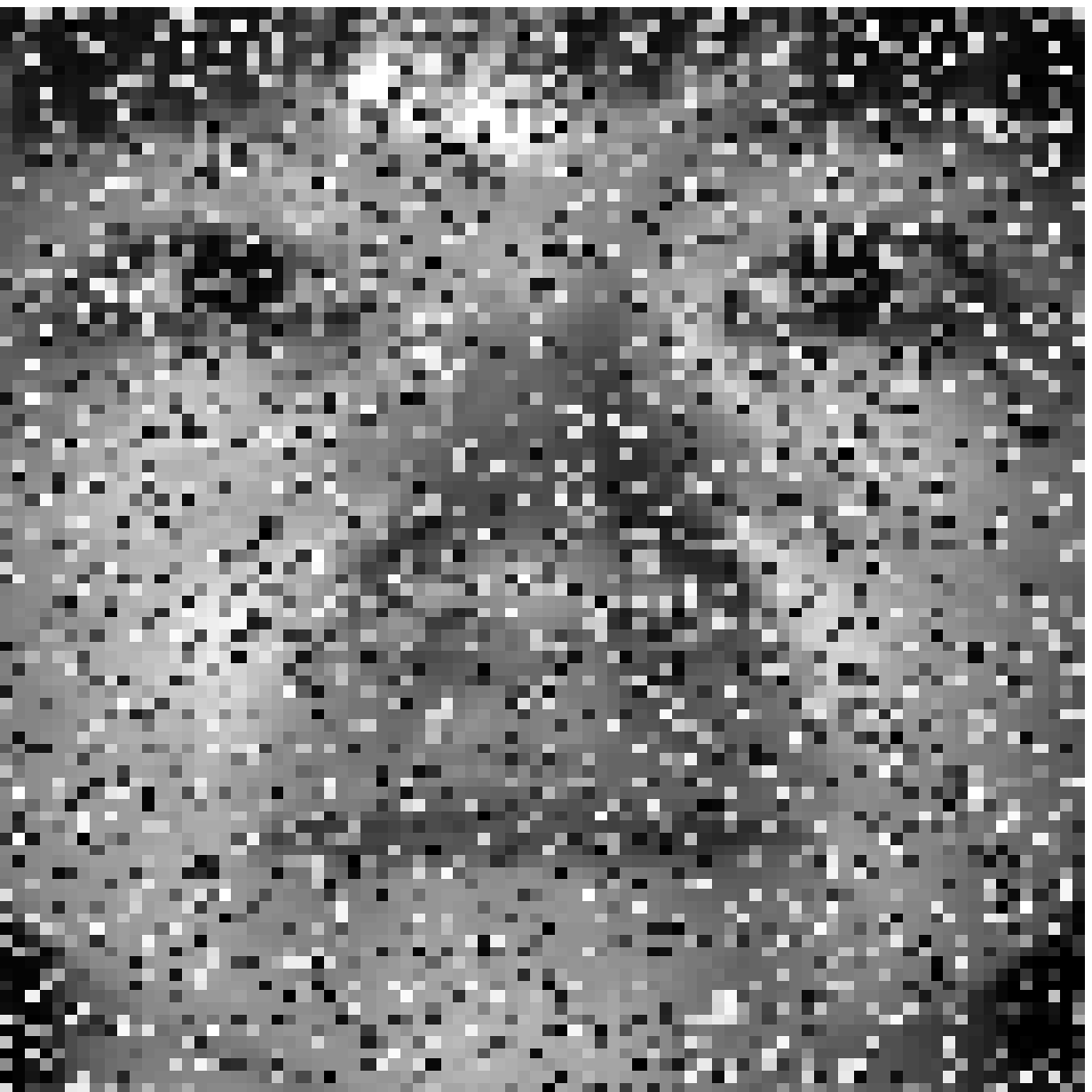}
}
\subfloat[$0.1827$]{%
\includegraphics[width=0.175\textwidth]{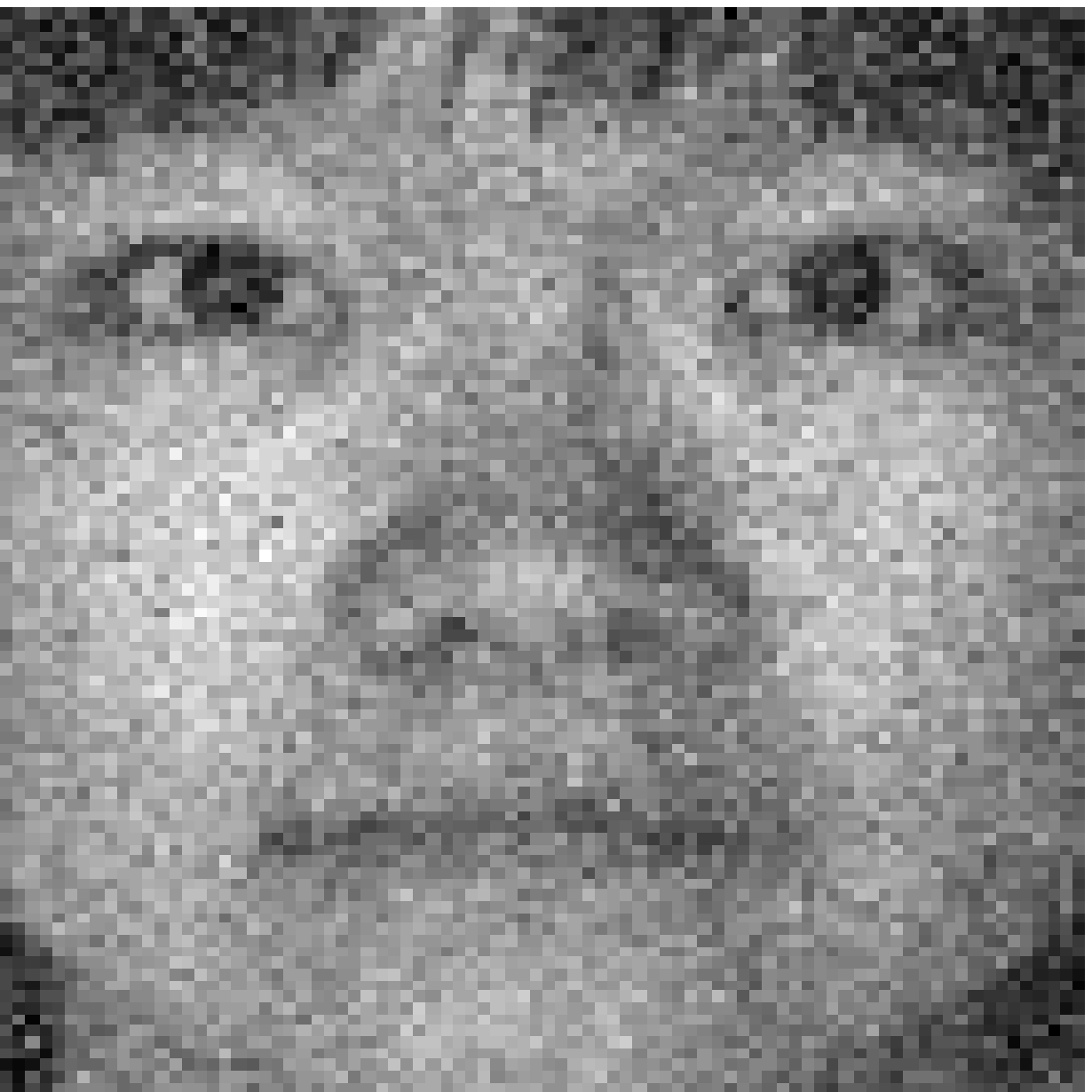}
}
\subfloat[$0.1517$]{%
\includegraphics[width=0.175\textwidth]{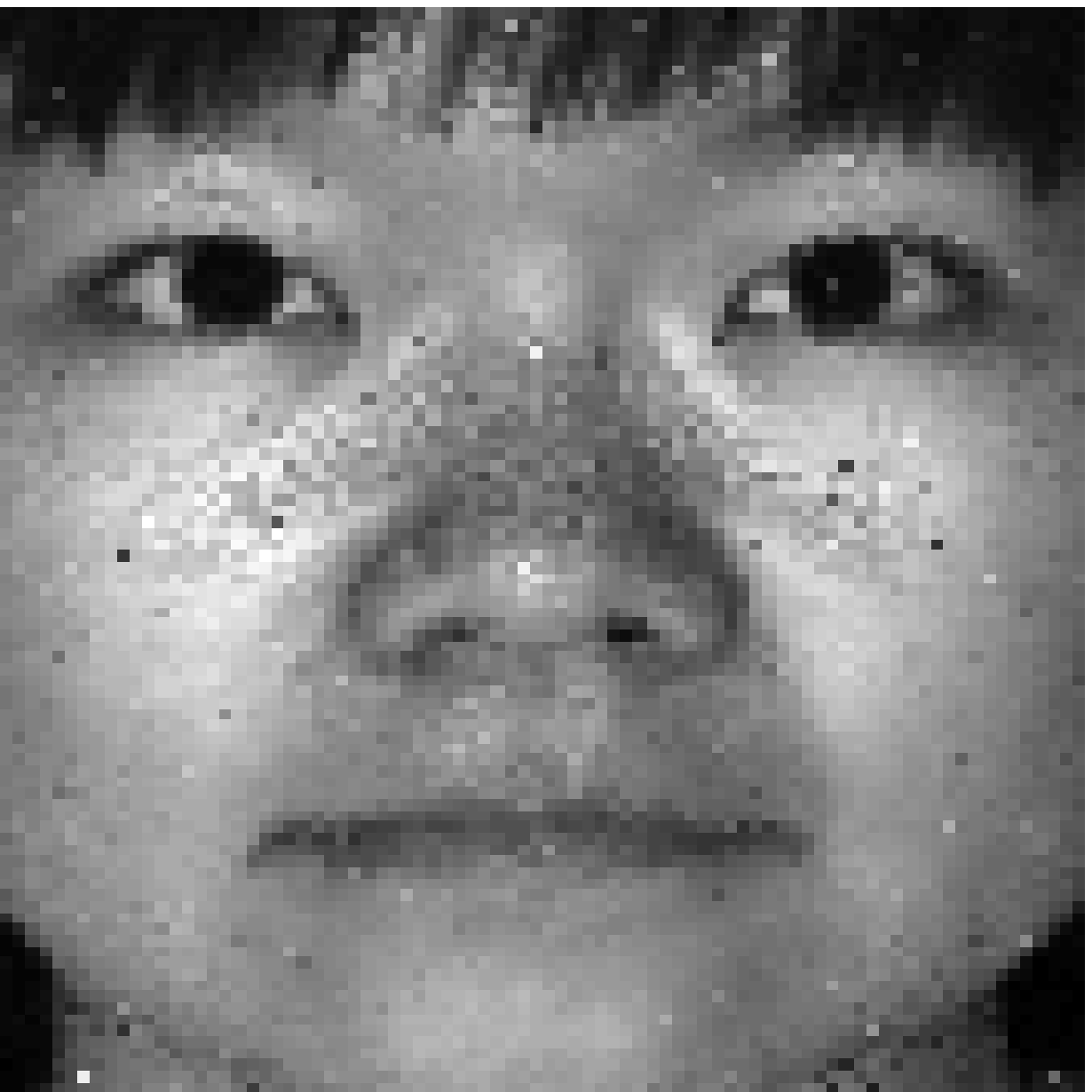}
} \\
\subfloat[$0$]{%
\includegraphics[width=0.175\textwidth]{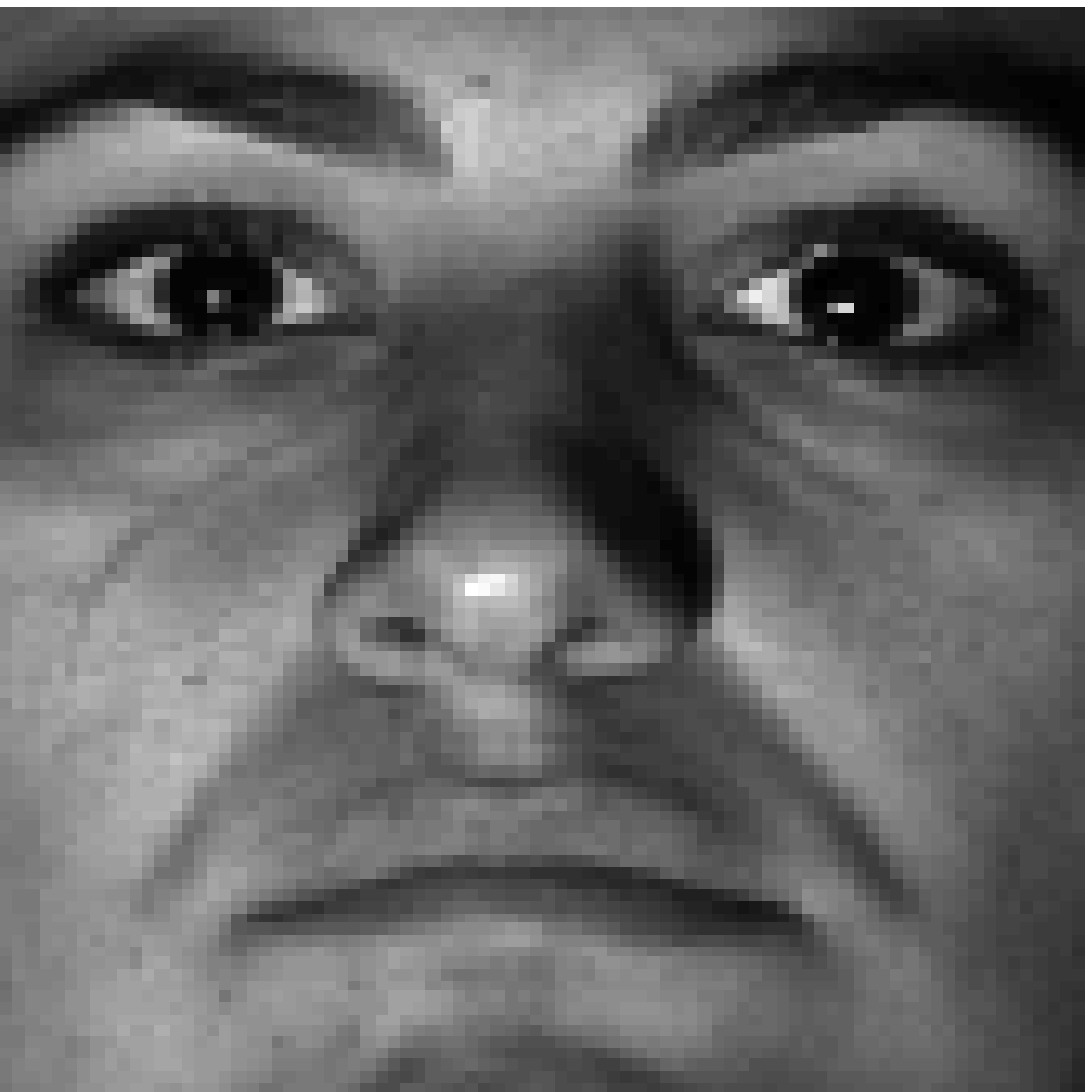}
}
\subfloat[$1.0304$]{%
\includegraphics[width=0.175\textwidth]{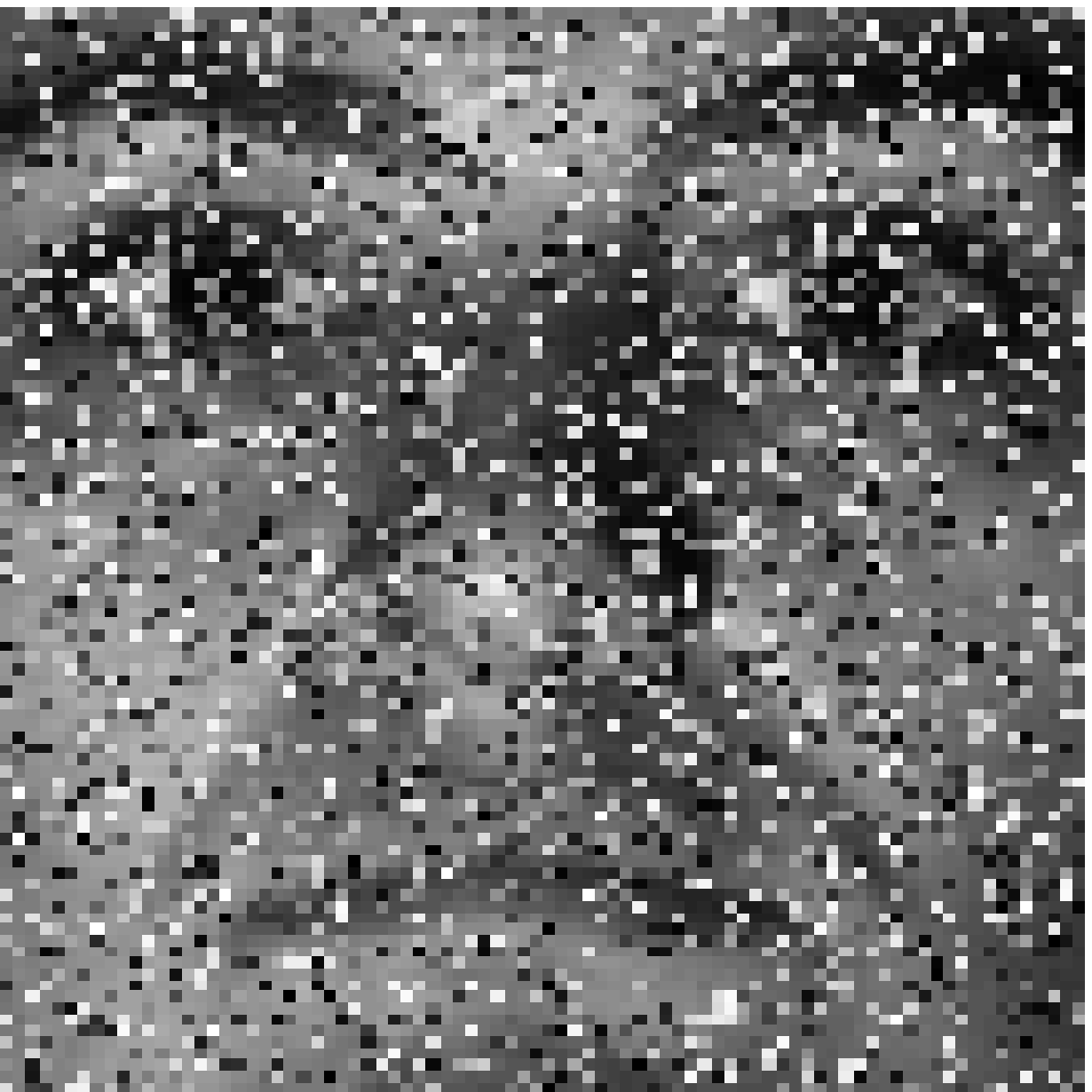}
}
\subfloat[$0.2013$]{%
\includegraphics[width=0.175\textwidth]{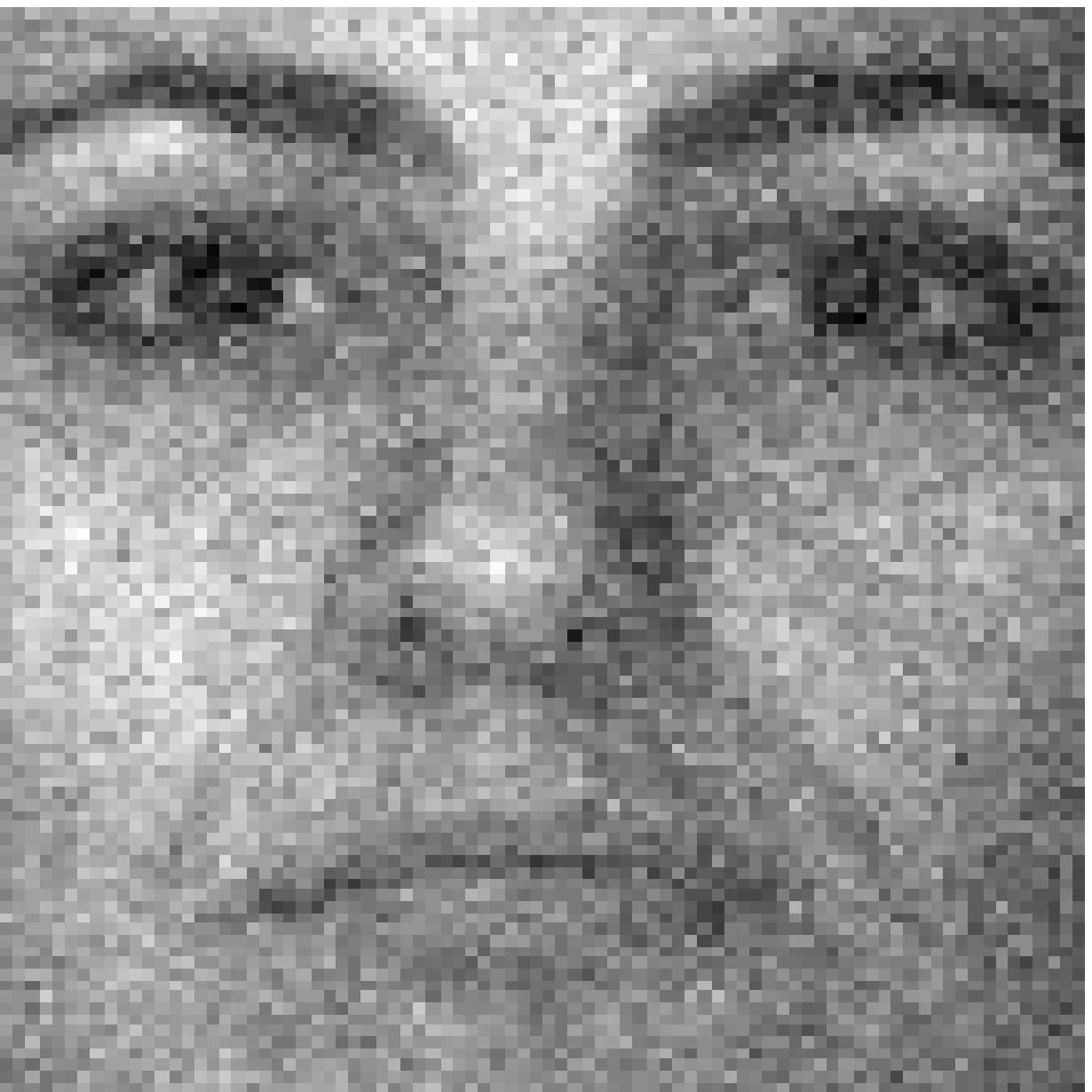}
}
\subfloat[$0.1865$]{%
\includegraphics[width=0.175\textwidth]{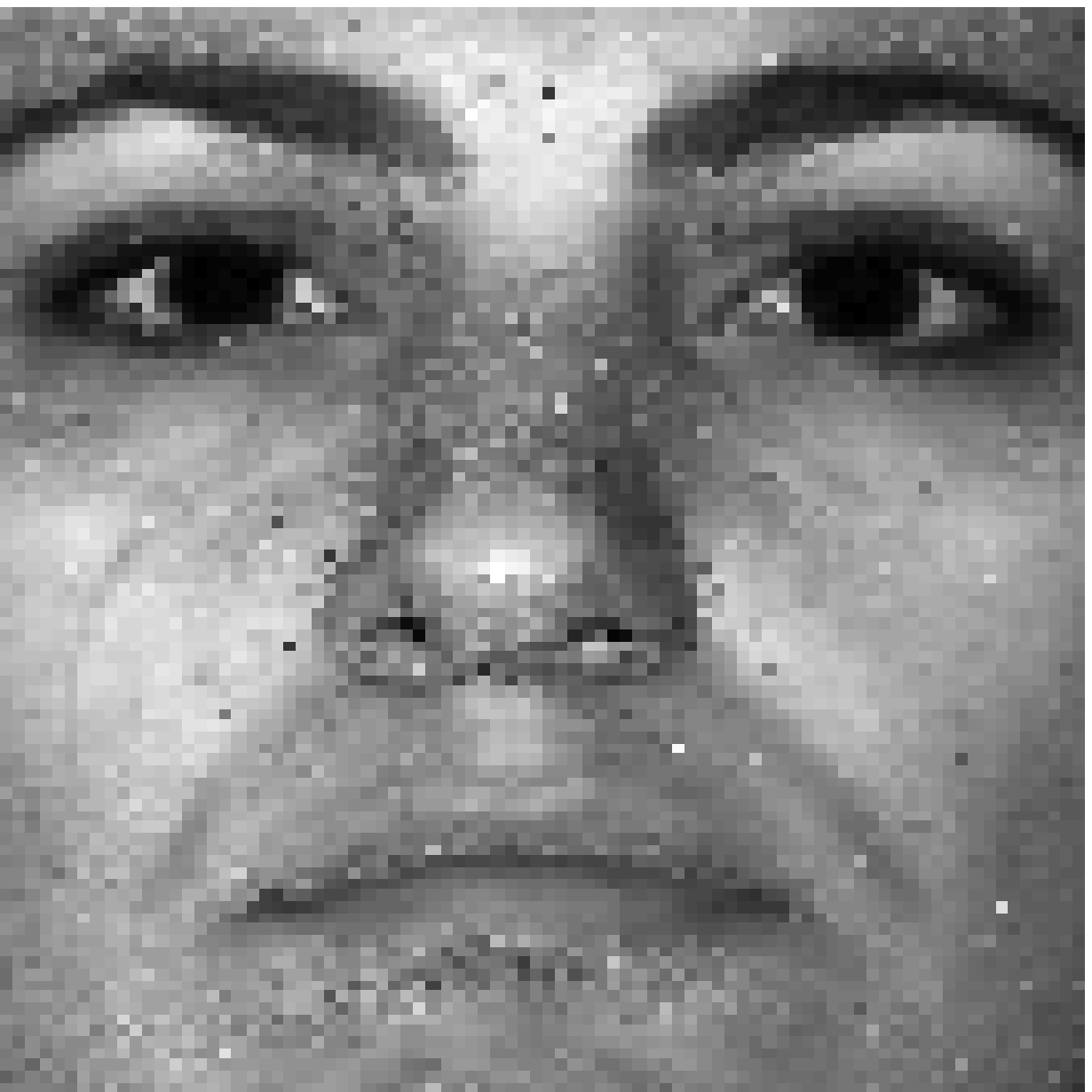}
} \\
\subfloat[$0$]{%
\includegraphics[width=0.175\textwidth]{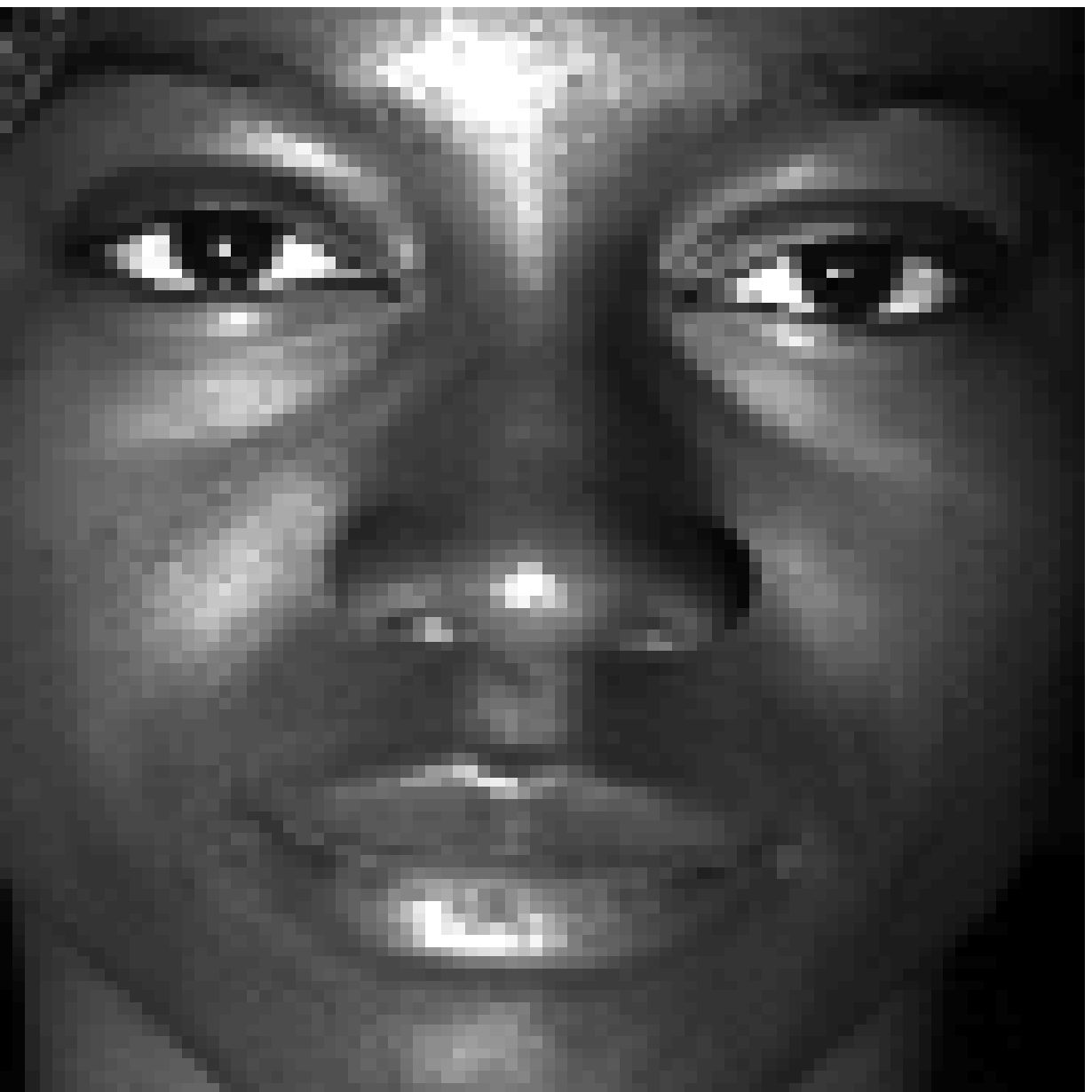}
}
\subfloat[$0.8192$]{%
\includegraphics[width=0.175\textwidth]{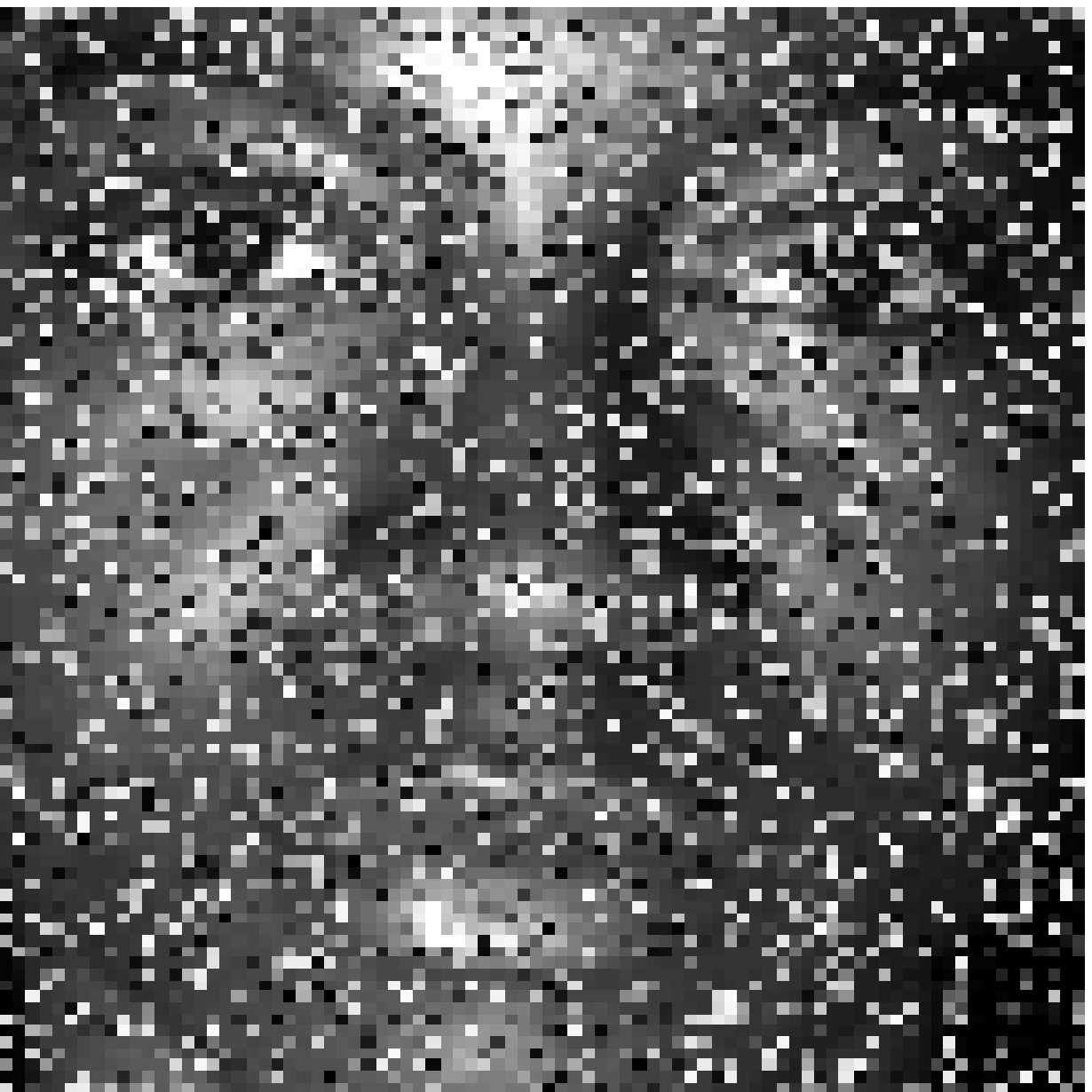}
}
\subfloat[$0.2910$]{%
\includegraphics[width=0.175\textwidth]{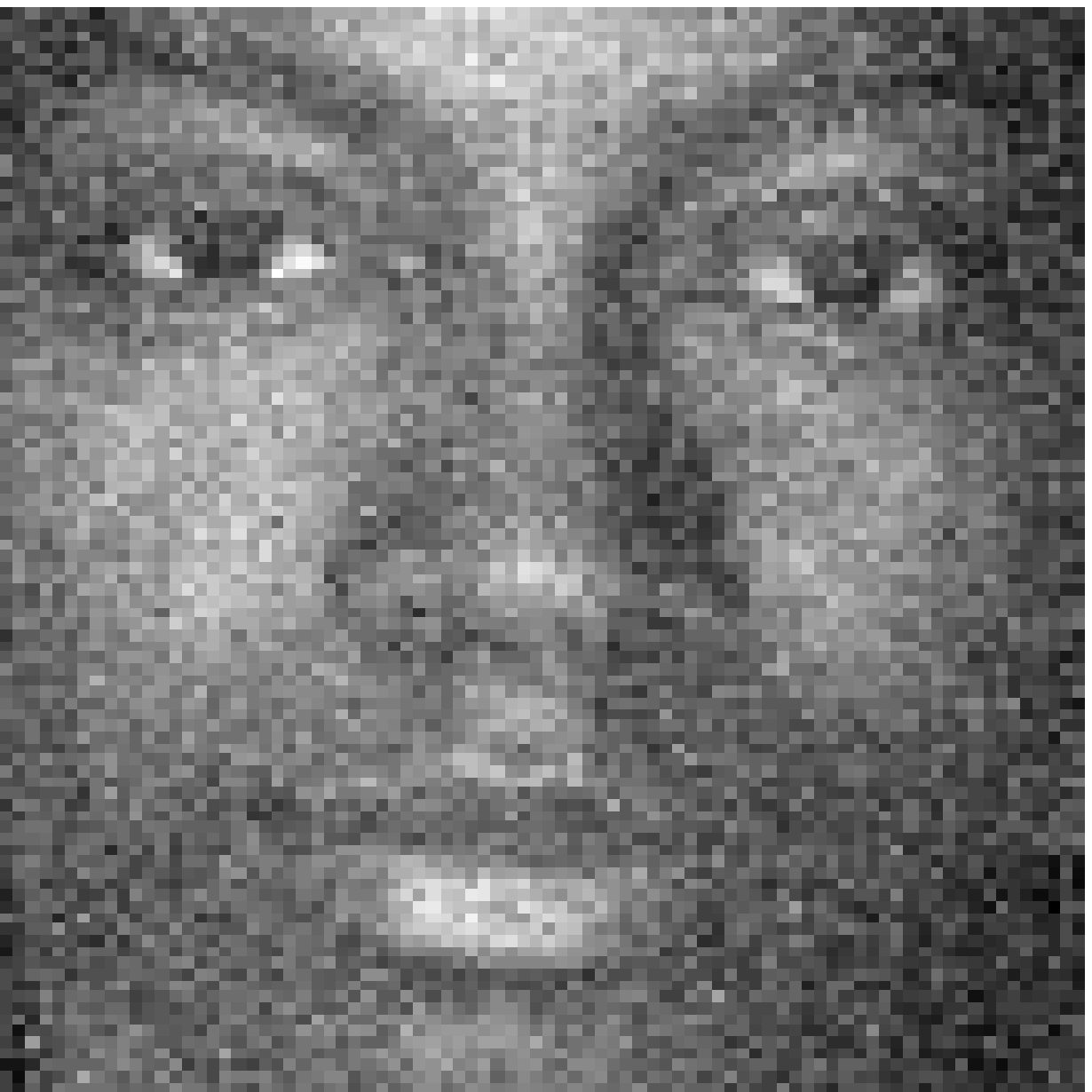}
}
\subfloat[$0.2832$]{%
\includegraphics[width=0.175\textwidth]{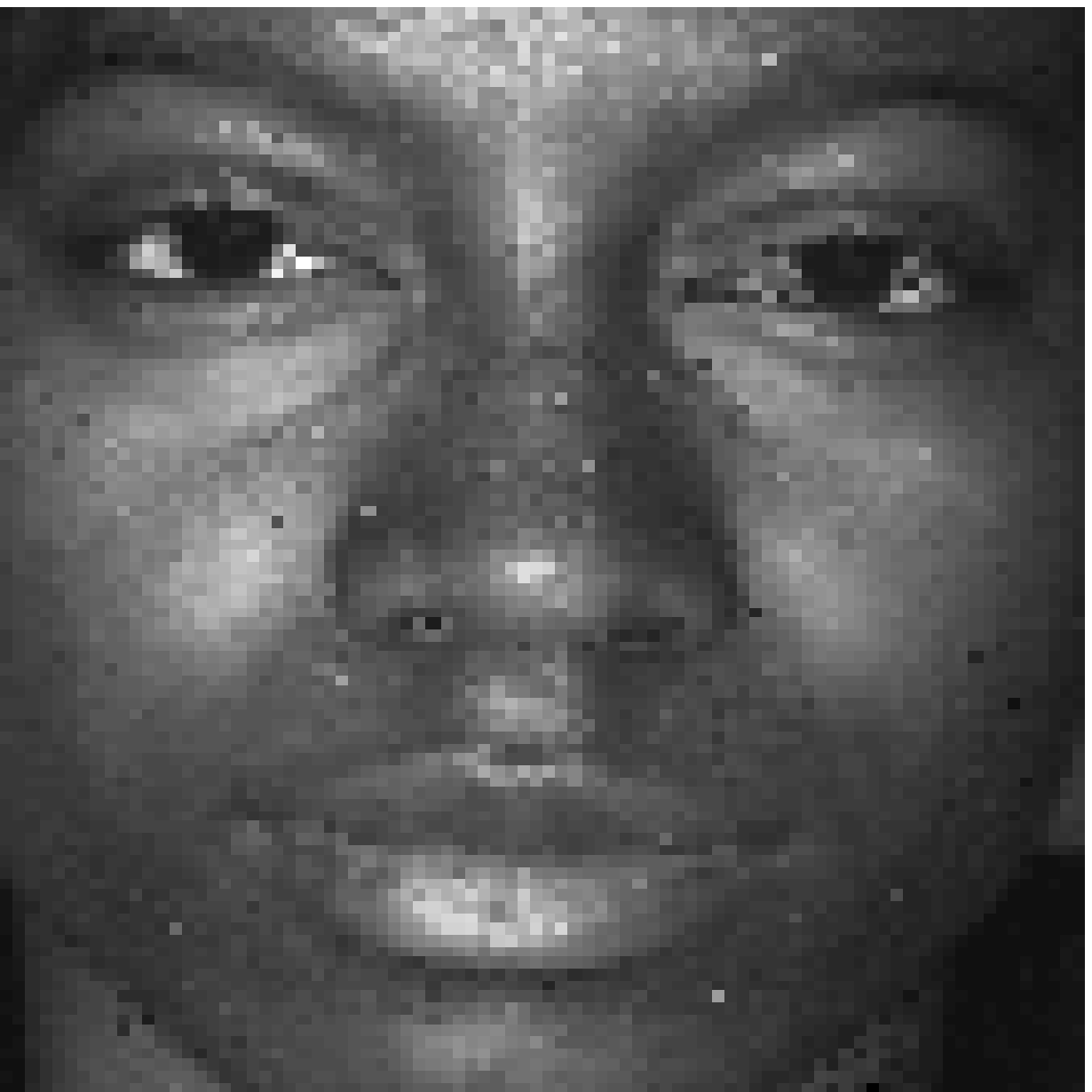}
} \\
\subfloat[$0$]{%
\includegraphics[width=0.175\textwidth]{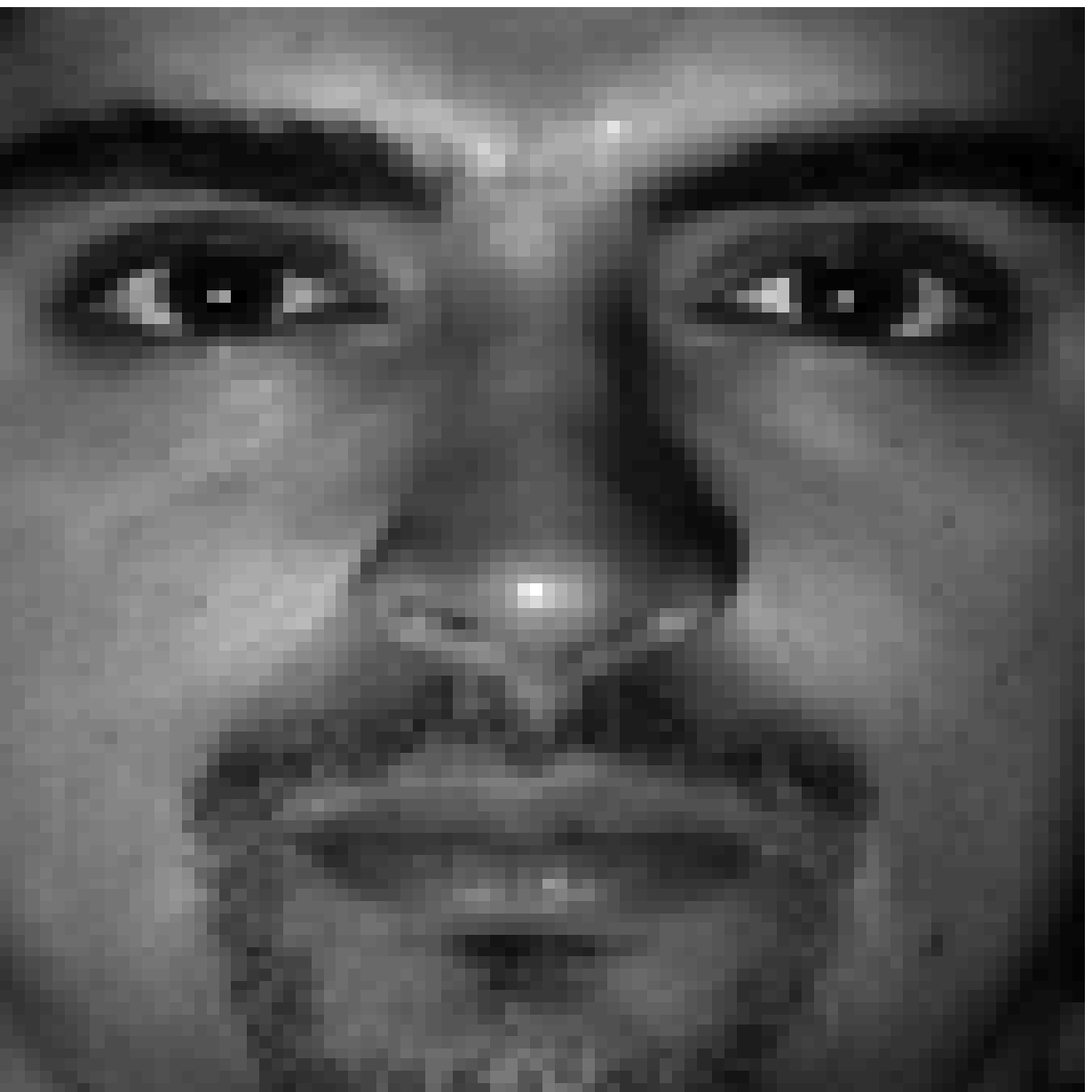}
}
\subfloat[$1.0032$]{%
\includegraphics[width=0.175\textwidth]{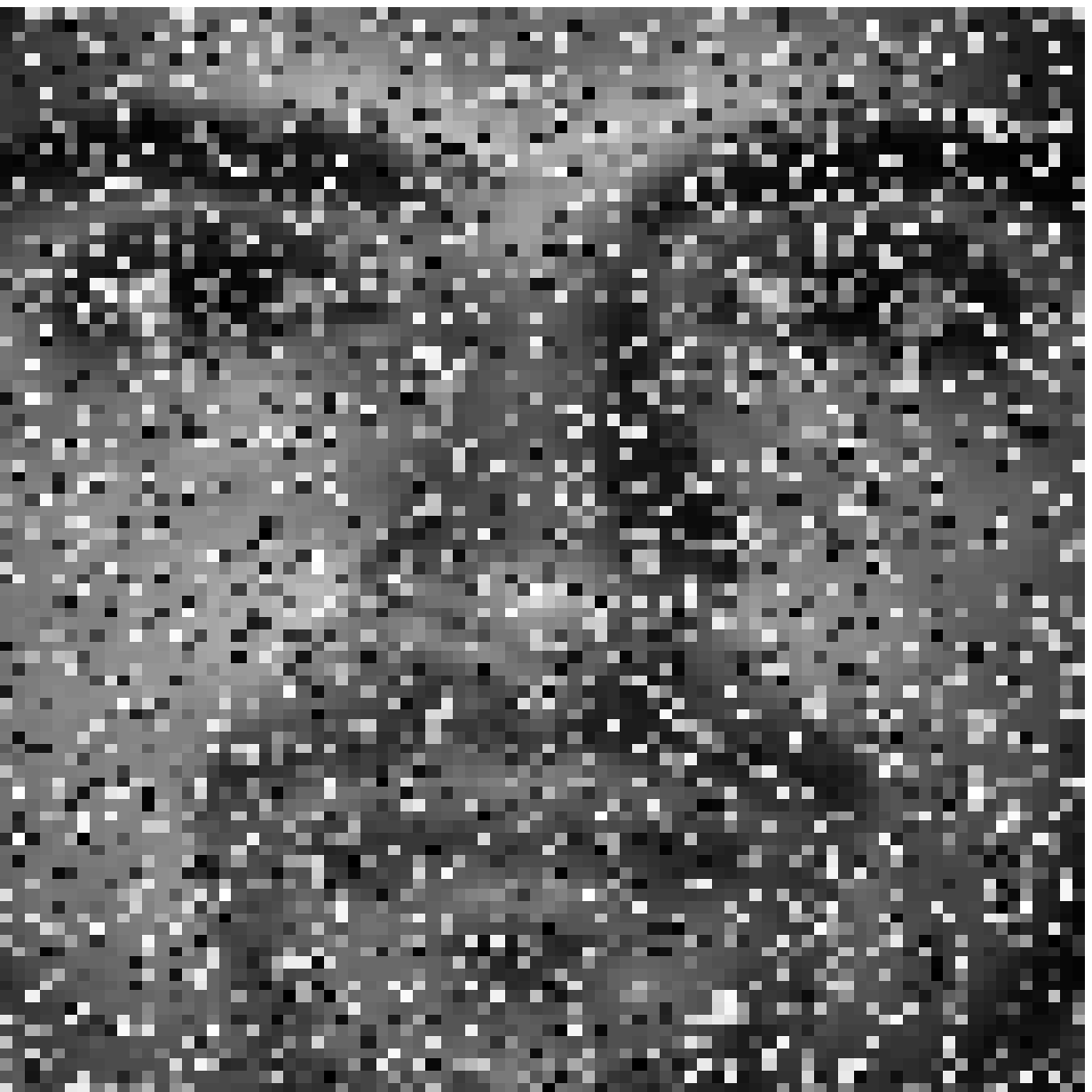}
}
\subfloat[$0.2158$]{%
\includegraphics[width=0.175\textwidth]{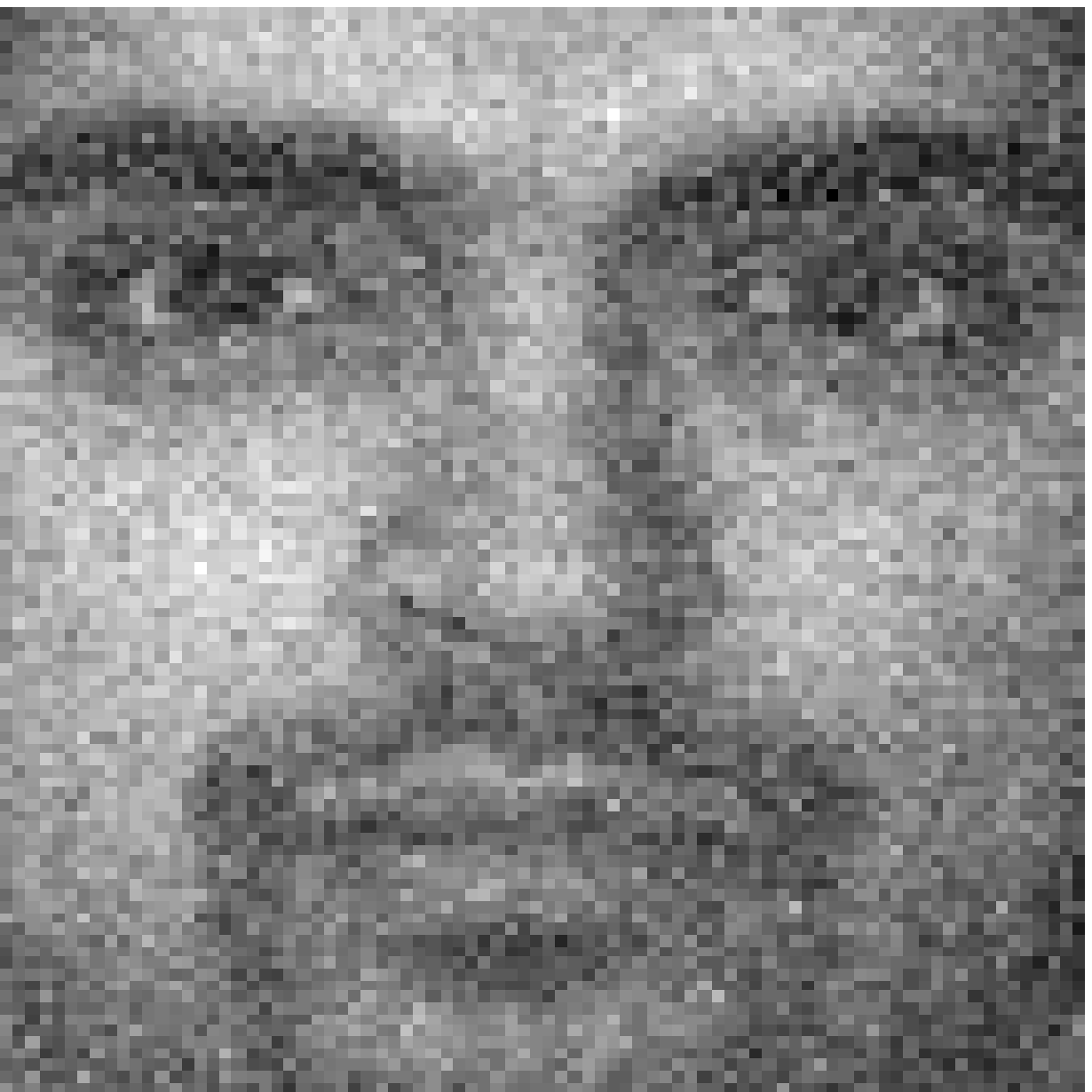}
}
\subfloat[$0.1835$]{%
\includegraphics[width=0.175\textwidth]{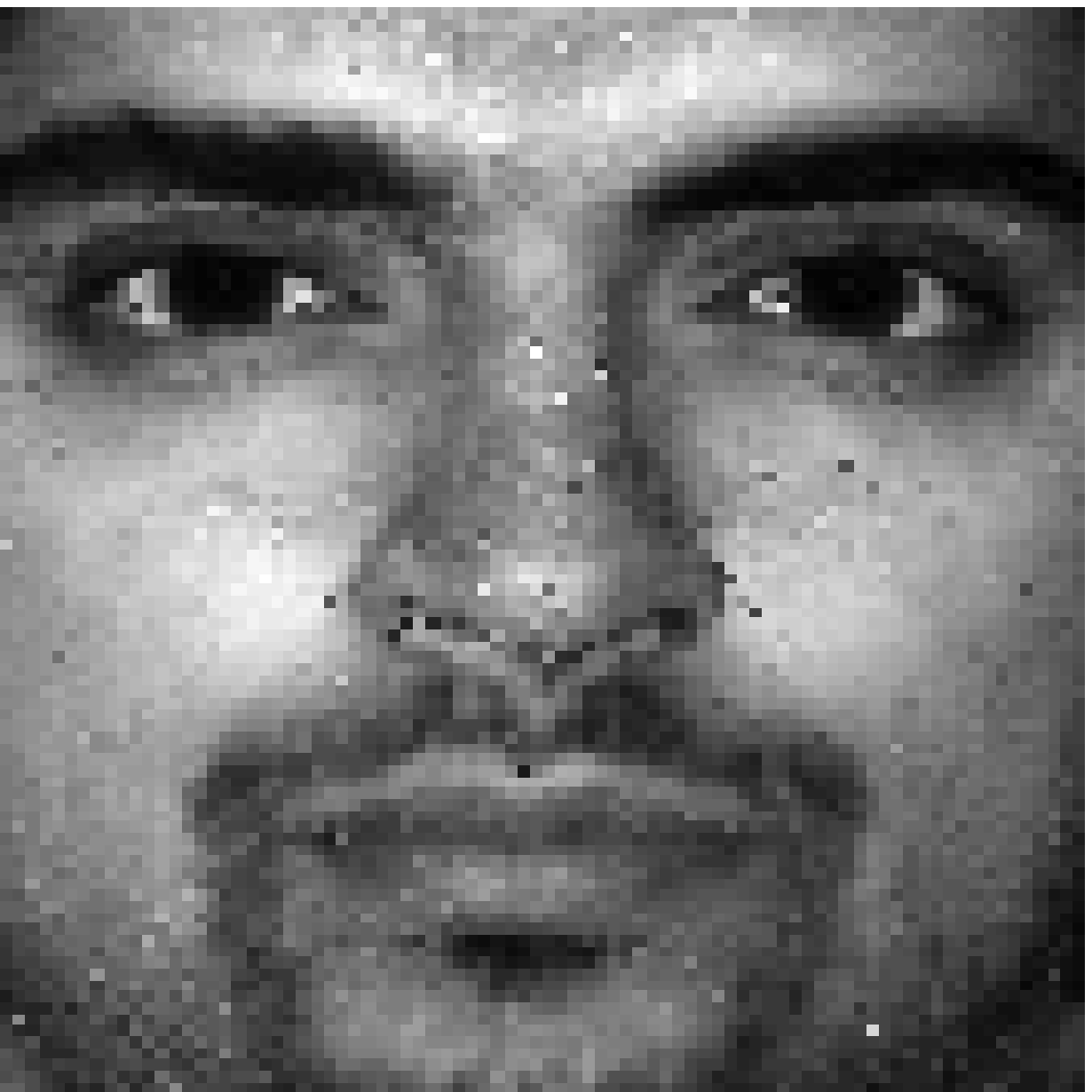}
}
\caption{Columns from left to right: original image, corrupted image, reconstruction by PPCA and reconstruction by Cauchy-PPCA. Columns two to four: below each image we quote its distance to the original image in the first column (lower is better), i.e. the quality of reconstruction.}
\label{fig:faces}
\end{figure*}

We applied the original PPCA \cite{Tipping1999} and the proposed Cauchy-PPCA on a task concerning the denoising of
images that have undergone pixel corruption. The aim here is to show the ease with which the proposed method can accommodate a change in the specification (i.e. change from Normal to Cauchy likelihood) and deliver a well-performing model.

The data $\bd{Y}$ are $2,414$ face images of $38$ individuals obtained from the Extended Yale B Database \cite{Georghiades2001}. There are $64$ images per individual under $9$ poses and $64$ illumination conditions. The images are grayscale images whose pixels have values between $0$ and $255$. 
We rescale the images to $96\times 84$ pixels. Hence $d=96\times 84=8064$ and we set $q=2$, i.e. both PCA schemes project the images to a latent space of dimension equal to $2$.
We corrupt $33.33\%$ of the pixels in each image by drawing a new value uniformly in the range $[0,\dots,255]$. 
For each individual, we use half of the corrupted images as the training set and the other half as the test set.

Fig.~\ref{fig:faces} presents results obtained on test images from $4$ individuals.
The figure shows from left to right the original and corrupted test image followed by the two reconstructions
obtained by PPCA and Cauchy-PPCA respectively. 
In order to quantify the quality of reconstruction, we use the following measure between the original and reconstructed images: $\|\bd{y}_{\mbox{\tiny orig}} - \bd{y}_{\mbox{\tiny rec}} \|^2 / \|\bd{y}_{\mbox{\tiny orig}} \|^2$. This measure is quoted below each image. 
The results in Fig.~\ref{fig:faces} evidently show that Cauchy-PPCA achieves better denoising levels than PPCA. In actual fact, in our numerical experiments we found that Cauchy-PPCA outperformed PPCA on all $38$ individuals. 

The present numerical experiment demonstrates the versatility of the proposed method in how it can easily extend PPCA to incorporate a Cauchy likelihood. This is achieved without exploiting any particular knowledge pertaining to the probabilistic specification of the model.

\subsection{Bayesian Inference for the Stochastic Model by Boore}
\label{sec:psha}

\begin{figure*}[t]
\centering
\includegraphics[width=0.6\textwidth]{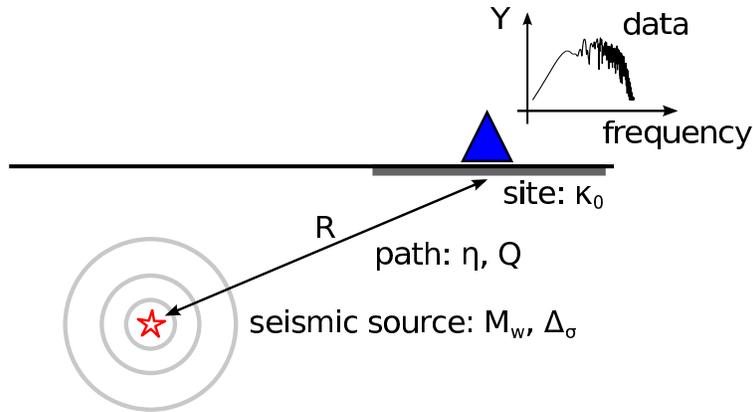}
\caption{Physical setting of seismic wave propagation from the source to the site of interest. Recorded at the site is the signal Fourier amplitude against frequency of ground motion.}
\label{fig:stochastic}
\end{figure*}

In this section we apply the proposed scheme on a geophysical
model called the stochastic model.
The stochastic model, due to Boore \cite{Boore2003},
is used to predict ground motion at a given site of interest
caused by an earthquake.
Ground motion is simply the shaking of the earth and it is
a fundamental quantity in estimating the seismic hazard
of structures.
From a physical point of view, the stochastic model describes the
radiated source spectrum and its amplitude changes in the frequency
domain due to wave propagation from the source to the site of interest. 
The inputs to the stochastic model are the distance $R$ of the site of
interest to the seismic source, the magnitude $M_w$ of the earthquake,
and the frequency $f$ of ground motion.
The stochastic model, in its simple form, has a parameter associated with
the seismic source known in the literature as stress parameter ($\Delta\sigma$),
two parameters associated with the path attenuation called geometrical spreading ($\eta$)
and quality factor ($Q$), and one
more parameter associated with the site called near-surface attenuation ($\kappa_0$).
All aforementioned parameters are bounded within a physical range.
In the case of multiple seismic sources, each source is associated with its own
distinct stress parameter.
The scalar output of the model $y$ is the mean Fourier amplitude of the ground motion.
The type of ground motion we consider here is acceleration. 
We denote the stochastic model as a function $y=g(M_w,R,f ; \bd{w})$, where
$\bd{w}=[\Delta\sigma_1,\dots,\Delta\sigma_E, \eta, Q, \kappa_0]$,
where $E$ is the number of seismic sources.
This situation is depicted in Fig. \ref{fig:stochastic}.
We refer the interested
reader to \cite{Boore2003} for more details.
Estimating the posterior uncertainty of the model parameters is important 
in seismic hazard analysis as the propagation of uncertainty in the 
parameters can have an impact on the estimated hazard curve \cite{Reiter}.
A discussion of how these posteriors can be utilised in 
later stages of seismic hazard analysis is beyond the scope of this work.

We specify the model by choosing the following densities:
\begin{itemize}
\item $\begin{fleqn}[0pt]\begin{aligned}[t] \mbox{Likelihood:\ }  & p(\bd{Y}|\mathcal{D},\bd{w}) = \\  &\prod_{n=1}^N \mathcal{N}(y_n | g({M_w}_n,R_n,f_n;\bd{w}),\sigma^2) \ . \end{aligned}\end{fleqn}$
\item Flat prior: $p(\bd{w}) \propto 1$\ .
\item Postulated posterior: $q(\bd{w}) = \mathcal{N}(\bd{w} | \bd{\mu},\bd{L}\bd{L}^T)$\ .
\end{itemize}
In contrast to the previous applications, here we choose a very flat prior. Ground motion data inputs are denoted by $\mathcal{D}$
and targets by \bd{Y}.

We performed experiments on a subset of the recently compiled RESORCE database \cite{Akkar2014}
which is a pan-European database of strong-motion recordings.
In particular we focused on data records originating from a station in 
the region of L'Aquilla for $E=8$ seismic sources.
Hence, the total number of free model parameters in \bd{w} is $11$.
We experimented with varying number of data records, $N\in\{100,200,500,1000\}$,
in order to test the robustness of the Laplace and the proposed approximation in scenarios
of limited data. Such situations arise in geophysical studies when data recordings are incomplete
due to distortions in frequencies caused by instrumentation errors.
The performance of Laplace and the proposed scheme was evaluated 
as the prediction error on  test sets. Both schemes
were run $10$ times, and each run involved a new random realisation
of the training and testing set. Parameter $S$ was set to $1000$ for all experiments in this section.
The predictive distribution for the proposed scheme 
was approximated
using a Monte Carlo estimate. We drew $200$ parameter samples
from the Gaussian fitted posterior $q$ and
estimated performance on the testing set as
the average of the mean squared error under each parameter sample.
The results are reported in Table \ref{tbl:results_ground_motion}.

\begin{table}[h]
\caption{Prediction error for ground motion problem (lower is better).}
  \label{tbl:results_ground_motion}
\centering
  \begin{tabular}{| c | c | c |}
    \hline
    $N$  & Laplace              & Proposed  \\
    \hline        
    100  & $0.8550 \pm  0.4008$ & $0.6040 \pm 0.0307$ \\ \hline
    200  & $0.7411 \pm  0.4925$ & $0.5776 \pm 0.0403$ \\ \hline
    500  & $0.6926 \pm  0.4790$ & $0.5496 \pm 0.0451$ \\ \hline
    1000 & $0.5395 \pm  0.0230$ & $0.5323 \pm 0.0273$ \\ \hline
  \end{tabular}
\end{table}

The results show that the proposed approximation fares better than Laplace, although
at $N=1000$ the performances are virtually identical.  For lower $N$, however, the 
Laplace approximation exhibits much higher variance than the proposed scheme.

\section{Discussion and Conclusion}
\label{sec:discussion}

We have presented a scheme for Bayesian variational inference that is applicable
in cases where the likelihood function renders more standard
approaches difficult. The scheme is conceptually simple as it
relies on a simple Monte Carlo average of the intractable
part of the variational lower bound, see Eq. (\ref{eq:lower_bounded_marg_likel}), and the re-introduction of
the variational parameters resulting in the objective
of Eq. (\ref{eq:approx_objective}). The scheme can thus be generally applied to other models where variational inference is difficult requiring only the gradients of the log-likelihood function with respect to the parameters. 

In the numerical experiments we have shown
that (a) the proposed scheme stands in close agreement with exact inference in Bayesian linear regression, (b) it performs up to par in classification tasks against methods that design bespoke model formulations, (c) it fares better than the Laplace approximation in a number of cases, and (d) it is very versatile and can be applied to a variety of problems.

Future work will address the relationship of our approach to
variational approaches \cite{FastGMRFLatent-08,EP-MarginalsGaussian-11} that provide alternative ways to compute improved Gaussian approximations to intractable posteriors relative to the Laplace approximation. 
Another aspect concerns ways to cope with very large problems that would require a large number of samples $S$ to obtain a sufficiently accurate approximation in Eq. \eqref{eq:sampled_term1_explicit}. A natural choice would be to turn the scheme in Algorithm \ref{proposed_scheme} into a recursive stochastic optimisation scheme \cite{Kushner2003} that employs \emph{small} sample sets computed at \emph{each} iteration, akin to stochastic gradient-based large-scale empirical risk minimisation \cite{Bottou2012}. These two approaches should not be confused, however. The latter employs subsampling of the \emph{data} $(\bd{X},\bd{Y})$ whereas our scheme generates samples based on current parameter estimates of the approximate posterior. Clearly, our scheme could incorporate sequential subsampling of the data as well. The problem of proving convergence of such an overall stochastic approximation approach in a suitable sense \cite{Kushner2003} seems to be open.

\section*{Acknowledgement}
The RESORCE database \cite{Akkar2014} was used in this work with the kind permission of the SIGMA project\footnote{\url{http://www.projet-sigma.com}}. N. Gianniotis was partially funded by the BMBF project ``Potsdam Research Cluster for Georisk Analysis, Environmental Change and Sustainability". C. Molkenthin and S. S. Bora were funded by the graduate research school GeoSim of the Geo.X initiative\footnote{\url{http://www.geo-x.net}}.

\bibliographystyle{plain}
\bibliography{main}

\begin{thebibliography}{10}

\bibitem{Akkar2014}
S.~Akkar, M.A. Sandikkaya, M.~Senyurt, A.~Azari~Sisi, B.~\"{O}. Ay,
  P.~Traversa, J.~Douglas, F.~Cotton, L.~Luzi, B.~Hernandez, and S.~Godey.
\newblock {R}eference {D}atabase for {S}eismic {G}round-{M}otion in {E}urope
  {(RESORCE)}.
\newblock {\em Bulletin of Earthquake Engineering}, 12(1):311--339, 2014.

\bibitem{Archambeau2006}
C{\'e}dric Archambeau, Nicolas Delannay, and Michel Verleysen.
\newblock {Robust Probabilistic Projections}.
\newblock In {\em {Proceedings of the 23rd International Conference on Machine
  Learning}}, pages 33--40. ACM, 2006.

\bibitem{Azzalini2005}
Adelchi Azzalini.
\newblock {T}he {S}kew-{N}ormal {D}istribution and {R}elated {M}ultivariate
  {F}amilies.
\newblock {\em Scandinavian Journal of Statistics}, 32(2):159--188, 2005.

\bibitem{Barber1998}
David Barber and Christopher~M. Bishop.
\newblock {E}nsemble {L}earning in {B}ayesian {N}eural {N}etworks.
\newblock In {\em Generalization in Neural Networks and Machine Learning},
  pages 215--237. Springer, 1998.

\bibitem{Beal2003}
M.~J. Beal.
\newblock {\em {Variational Algorithms for Approximate Bayesian Inference}}.
\newblock PhD thesis, Gatsby Computational Neuroscience Unit, University
  College London, 2003.

\bibitem{Bishop}
C.~M. Bishop.
\newblock {\em {P}attern {R}ecognition and {M}achine {L}earning}.
\newblock Springer, 2006.

\bibitem{Boore2003}
D.~M. Boore.
\newblock {S}imulation of {G}round {M}otion {U}sing the {S}tochastic {M}ethod.
\newblock {\em Pure and Applied Geophysics}, 160(3-4):635--676, 2003.

\bibitem{Bottou2012}
L.~Bottou.
\newblock {S}tochastic {G}radient {T}ricks.
\newblock In {\em Neural Networks, Tricks of the Trade, Reloaded}, volume 7700
  of {\em LNCS}. Springer, 2012.

\bibitem{EP-MarginalsGaussian-11}
B.~Cseke and T.~Heskes.
\newblock {A}pproximate {M}arginals in {L}atent {G}aussian {M}odels.
\newblock {\em Journal of Machine Learning Research}, 12:417--454, 2011.

\bibitem{Georghiades2001}
Athinodoros~S. Georghiades, Peter~N. Belhumeur, and David Kriegman.
\newblock {From Few to Many: Illumination Cone Models for Face Recognition
  under Variable Lighting and Pose}.
\newblock {\em Pattern Analysis and Machine Intelligence, IEEE Transactions
  on}, 23(6):643--660, 2001.

\bibitem{Jaakkola2000}
Tommi Jaakkola and Michael Jordan.
\newblock {Bayesian Parameter Estimation via Variational Methods}.
\newblock {\em Statistics and Computing}, 10(1):25--37, 2000.

\bibitem{Kushner2003}
H.J. Kushner and G.G. Yin.
\newblock {\em {S}tochastic {A}pproximation and {R}ecursive {A}lgorithms and
  {A}pplications}.
\newblock Springer, 2nd edition, 2003.

\bibitem{Moller1993}
Martin~Fodslette M{\o}ller.
\newblock {A Scaled Conjugate Gradient Algorithm for Fast Supervised Learning}.
\newblock {\em Neural networks}, 6(4):525--533, 1993.

\bibitem{VariationalGaussianApprox-09}
M.~Opper and C.~Archambeau.
\newblock {T}he {V}ariational {G}aussian {A}pproximation {R}evisited.
\newblock {\em Neural Computation}, 21:786--792, 2009.

\bibitem{MatrixCookbook}
Kaare~B. Petersen and Michael~S. Pedersen.
\newblock {T}he {M}atrix {C}ookbook, November 15 2012.

\bibitem{Psorakis2010}
Ioannis Psorakis, Theodoros Damoulas, and Mark~A. Girolami.
\newblock {Multiclass Relevance Vector Machines: Sparsity and Accuracy}.
\newblock {\em IEEE Transactions on Neural Networks}, 21(10):1588--1598, 2010.

\bibitem{Reiter}
L.~Reiter.
\newblock {\em {E}arthquake {H}azard {A}nalysis: {I}ssues and {I}nsights}.
\newblock Colombia University Press, 1991.

\bibitem{FastGMRFLatent-08}
H.~Rue, S.~Martino, and N.~Chopin.
\newblock {A}pproximate {B}ayesian {I}nference for {L}atent {G}aussian {M}odels
  by {U}sing {I}ntegrated {N}ested {L}aplace {A}pproximations.
\newblock {\em Journal of the Royal Statistical Society: Series B},
  71(2):319--392, 2009.

\bibitem{Tipping2001}
Michael~E Tipping.
\newblock {Sparse Bayesian Learning and the Relevance Vector Machine}.
\newblock {\em Journal of Machine Learning Research}, 1:211--244, 2001.

\bibitem{Tipping1999}
Michael~E. Tipping and Christopher~M. Bishop.
\newblock {Probabilistic Principal Component Analysis}.
\newblock {\em Journal of the Royal Statistical Society: Series B},
  61(3):611--622, 1999.

\bibitem{Tzikas2008}
D.G. Tzikas, C.L. Likas, and N.P. Galatsanos.
\newblock {The Variational Approximation for Bayesian inference}.
\newblock {\em Signal Processing Magazine, IEEE}, 25(6):131--146, 2008.

\bibitem{Xie2015}
Pengtao Xie and Eric~P. Xing.
\newblock {Cauchy Principal Component Analysis}.
\newblock {\em CoRR}, abs/1412.6506, 2014.

\end{thebibliography}
\end{document}